%% file: main.tex
\documentclass{article}

\usepackage{iclr2027_conference,times}

\usepackage[utf8]{inputenc}
\usepackage[T1]{fontenc}
\usepackage{hyperref}
\usepackage{url}
\usepackage{array}
\usepackage{booktabs}
\usepackage{siunitx}
\usepackage{amsfonts}
\usepackage{amsmath}
\usepackage{amsthm}
\usepackage{amssymb}
\usepackage{nicefrac}
\usepackage{microtype}
\usepackage{graphicx}
\usepackage{placeins}

\theoremstyle{definition}

\input{longmemeval_chatbot_integration_draft}
\input{longmemeval_chat_suffix_disclosure_macros}

\title{Can an AI Assistant Really Forget?\\
Auditable Deletion from\\
Addressable Memory\thanks{Code and evidence: \url{https://github.com/VyLabs-AI/gemmasv}}}

\author{Vishwajith Ramesh \\
Vy Labs, Inc. \\
\texttt{vish@vylabs.ai}}

\iclrfinalcopy  

\hypersetup{hidelinks,pdfauthor={Vishwajith Ramesh},pdftitle={Can an AI Assistant Really Forget? Auditable Deletion from Addressable Memory},pdfsubject={Public preprint}}


\begin{document}

\maketitle
\raggedbottom
\lhead{Public preprint}

\begin{abstract}
\input{abstract_gemma}
\end{abstract}

\input{body_gemma}

\bibliographystyle{iclr2027_conference}
\bibliography{references}

\appendix
\input{appendix_gemma}

\end{document}

%% file: longmemeval_chatbot_integration_draft.tex
\ifdefined\LMChatTemplateSyntaxTest
  \providecommand{\LMChatMacrosGeneratedFromCompact}{true}
  \providecommand{\LMChatCompactSchema}
    {gemma-sv-longmemeval-chat-geometry-methods-compact16-v1}
  \providecommand{\LMChatValidationStatus}{validated-complete-all16}
  \providecommand{\LMChatBehavioralResultStatus}{validated-positive}
  \providecommand{\LMChatCertificateStatus}{validated-pass}
  \providecommand{\LMChatManuscriptConditionsComplete}{true}
  \providecommand{\LMChatOnlyExcludedFailuresReconciled}{true}
  \providecommand{\LMChatContainsSourceText}{false}
  \providecommand{\LMChatOfficialLeaderboardScore}{false}
  \providecommand{\LMChatNoOutputBasedReplacement}{true}
  \providecommand{\LMChatGeometryCorrectionOutputBlind}{true}
  \providecommand{\LMChatCoreBindingsUnchanged}{true}

  \providecommand{\LMChatFrozenN}{16}
  \providecommand{\LMChatAttemptedN}{16}
  \providecommand{\LMChatTerminalN}{16}
  \providecommand{\LMChatNotAttemptedN}{0}
  \providecommand{\LMChatTargetAdmittedN}{10}
  \providecommand{\LMChatRetainedAvailableN}{16}
  \providecommand{\LMChatJointN}{10}
  \providecommand{\LMChatSourceImmutableN}{16}
  \providecommand{\LMChatRawOmissionCompleteN}{16}
  \providecommand{\LMChatExactPolicyCompleteN}{16}
  \providecommand{\LMChatRefitCompleteN}{16}
  \providecommand{\LMChatFPThirtyTwoCompleteN}{16}
  \providecommand{\LMChatDecayCompleteN}{16}
  \providecommand{\LMChatPromptCompleteN}{16}
  \providecommand{\LMChatCertificateRecordN}{16}
  \providecommand{\LMChatCertificateProbeN}{32}
  \providecommand{\LMChatIncrementalN}{0}
  \providecommand{\LMChatRefitFallbackN}{16}
  \providecommand{\LMChatRawFallbackN}{0}

  \providecommand{\LMChatMetricPlaceholder}{\mathrm{compact16}}
  \providecommand{\LMChatPresentSuppression}{\LMChatMetricPlaceholder}
  \providecommand{\LMChatRawSuppression}{\LMChatMetricPlaceholder}
  \providecommand{\LMChatExactSuppression}{\LMChatMetricPlaceholder}
  \providecommand{\LMChatRefitSuppression}{\LMChatMetricPlaceholder}
  \providecommand{\LMChatFPThirtyTwoSuppression}{\LMChatMetricPlaceholder}
  \providecommand{\LMChatDecaySuppression}{\LMChatMetricPlaceholder}
  \providecommand{\LMChatPromptSuppression}{\LMChatMetricPlaceholder}
  \providecommand{\LMChatPresentTargetRawKL}{\LMChatMetricPlaceholder}
  \providecommand{\LMChatRawTargetRawKL}{\LMChatMetricPlaceholder}
  \providecommand{\LMChatExactTargetRawKL}{\LMChatMetricPlaceholder}
  \providecommand{\LMChatRefitTargetRawKL}{\LMChatMetricPlaceholder}
  \providecommand{\LMChatFPThirtyTwoTargetRawKL}{\LMChatMetricPlaceholder}
  \providecommand{\LMChatDecayTargetRawKL}{\LMChatMetricPlaceholder}
  \providecommand{\LMChatPromptTargetRawKL}{\LMChatMetricPlaceholder}
  \providecommand{\LMChatPresentRetainedRawKL}{\LMChatMetricPlaceholder}
  \providecommand{\LMChatRawRetainedRawKL}{\LMChatMetricPlaceholder}
  \providecommand{\LMChatExactRetainedRawKL}{\LMChatMetricPlaceholder}
  \providecommand{\LMChatRefitRetainedRawKL}{\LMChatMetricPlaceholder}
  \providecommand{\LMChatFPThirtyTwoRetainedRawKL}{\LMChatMetricPlaceholder}
  \providecommand{\LMChatDecayRetainedRawKL}{\LMChatMetricPlaceholder}
  \providecommand{\LMChatPromptRetainedRawKL}{\LMChatMetricPlaceholder}
  
  \providecommand{\LMChatCertificateMeanKL}{\LMChatMetricPlaceholder}
  \providecommand{\LMChatCertificateMaxKL}{\LMChatMetricPlaceholder}

  \providecommand{\LMChatExactBelowRawFloor}{\LMChatMetricPlaceholder}
  \providecommand{\LMChatExactAtOrBelowRawN}{6}
  
  \providecommand{\LMChatCacheIncompatibleN}{16}
  \providecommand{\LMChatRawReportFailureN}{16}
  \providecommand{\LMChatPrimaryFailureN}{0}
  \providecommand{\LMChatCacheDeleteShiftClassification}
    {incompatible-diagnostic}
  \providecommand{\LMChatCacheDeleteShiftFailedN}{16}

\else
  \InputIfFileExists{longmemeval_chatbot_compact16_macros.tex}{}{C
    \PackageError{longmemeval-chatbot-integration}
      {Validated compact16 macro file is missing}
      {Do not use partial results. Generate the macro file only from the
       validated compact all-16 report.}%
  }
\fi

\InputIfFileExists{longmemeval_chat_v3_decoded_macros.tex}{}{C
  \PackageError{longmemeval-chatbot-integration}

    {Validated v3 decoded macro file is missing}
    {Regenerate it from the source-free v3 decoded and census statistics.}%
}

\makeatletter
\newcommand{\LMChatRequireMacro}[1]{%
  \@ifundefined{#1}{%
    \PackageError{longmemeval-chatbot-integration}
      {Required compact16 macro #1 is undefined}
      {Regenerate macros from the validated compact all-16 report.}%
  }{}%
}
\makeatother

\newcommand{\LMChatExpectText}[3]{%
  \begingroup
  \edef\LMChatActualValue{#1}%
  \edef\LMChatExpectedValue{#2}%
  \ifx\LMChatActualValue\LMChatExpectedValue
  \else
    \PackageError{longmemeval-chatbot-integration}
      {Compact16 guard failed for #3}
      {Expected `#2'. Refuse manuscript integration.}%
  \fi
  \endgroup
}

\newcommand{\LMChatExpectCount}[3]{%
  \ifnum\numexpr#1\relax=#2\relax
  \else
    \PackageError{longmemeval-chatbot-integration}
      {Compact16 count guard failed for #3}
      {Expected #2. Refuse partial or mismatched manuscript integration.}%
  \fi
}

\LMChatRequireMacro{LMChatMacrosGeneratedFromCompact}
\LMChatRequireMacro{LMChatCompactSchema}
\LMChatRequireMacro{LMChatValidationStatus}
\LMChatRequireMacro{LMChatBehavioralResultStatus}
\LMChatRequireMacro{LMChatCertificateStatus}
\LMChatRequireMacro{LMChatManuscriptConditionsComplete}
\LMChatRequireMacro{LMChatOnlyExcludedFailuresReconciled}
\LMChatRequireMacro{LMChatContainsSourceText}
\LMChatRequireMacro{LMChatOfficialLeaderboardScore}
\LMChatRequireMacro{LMChatNoOutputBasedReplacement}
\LMChatRequireMacro{LMChatGeometryCorrectionOutputBlind}
\LMChatRequireMacro{LMChatCoreBindingsUnchanged}
\LMChatRequireMacro{LMChatFrozenN}
\LMChatRequireMacro{LMChatAttemptedN}
\LMChatRequireMacro{LMChatTerminalN}
\LMChatRequireMacro{LMChatNotAttemptedN}
\LMChatRequireMacro{LMChatTargetAdmittedN}
\LMChatRequireMacro{LMChatRetainedAvailableN}
\LMChatRequireMacro{LMChatJointN}
\LMChatRequireMacro{LMChatSourceImmutableN}
\LMChatRequireMacro{LMChatRawOmissionCompleteN}
\LMChatRequireMacro{LMChatExactPolicyCompleteN}
\LMChatRequireMacro{LMChatRefitCompleteN}
\LMChatRequireMacro{LMChatFPThirtyTwoCompleteN}
\LMChatRequireMacro{LMChatDecayCompleteN}
\LMChatRequireMacro{LMChatPromptCompleteN}
\LMChatRequireMacro{LMChatCertificateRecordN}
\LMChatRequireMacro{LMChatCertificateProbeN}
\LMChatRequireMacro{LMChatIncrementalN}
\LMChatRequireMacro{LMChatRefitFallbackN}
\LMChatRequireMacro{LMChatRawFallbackN}
\LMChatRequireMacro{LMChatPresentSuppression}
\LMChatRequireMacro{LMChatRawSuppression}
\LMChatRequireMacro{LMChatExactSuppression}
\LMChatRequireMacro{LMChatRefitSuppression}
\LMChatRequireMacro{LMChatFPThirtyTwoSuppression}
\LMChatRequireMacro{LMChatDecaySuppression}
\LMChatRequireMacro{LMChatPromptSuppression}
\LMChatRequireMacro{LMChatPresentTargetRawKL}
\LMChatRequireMacro{LMChatRawTargetRawKL}
\LMChatRequireMacro{LMChatExactTargetRawKL}
\LMChatRequireMacro{LMChatRefitTargetRawKL}
\LMChatRequireMacro{LMChatFPThirtyTwoTargetRawKL}
\LMChatRequireMacro{LMChatDecayTargetRawKL}
\LMChatRequireMacro{LMChatPromptTargetRawKL}
\LMChatRequireMacro{LMChatPresentRetainedRawKL}
\LMChatRequireMacro{LMChatRawRetainedRawKL}
\LMChatRequireMacro{LMChatExactRetainedRawKL}
\LMChatRequireMacro{LMChatRefitRetainedRawKL}
\LMChatRequireMacro{LMChatFPThirtyTwoRetainedRawKL}
\LMChatRequireMacro{LMChatDecayRetainedRawKL}
\LMChatRequireMacro{LMChatPromptRetainedRawKL}
\LMChatRequireMacro{LMChatExactRetainedRank}
\LMChatRequireMacro{LMChatCertificateMeanKL}
\LMChatRequireMacro{LMChatCertificateMaxKL}
\LMChatRequireMacro{LMChatExactSuppressionMin}
\LMChatRequireMacro{LMChatExactSuppressionMax}
\LMChatRequireMacro{LMChatExactBelowRawFloor}
\LMChatRequireMacro{LMChatExactAtOrBelowRawN}
\LMChatRequireMacro{LMChatExactMeanAbsRetainedDrift}
\LMChatRequireMacro{LMChatCacheIncompatibleN}
\LMChatRequireMacro{LMChatRawReportFailureN}
\LMChatRequireMacro{LMChatPrimaryFailureN}
\LMChatRequireMacro{LMChatCacheDeleteShiftClassification}
\LMChatRequireMacro{LMChatCacheDeleteShiftFailedN}
\LMChatRequireMacro{LMChatFullReportSHA}
\LMChatRequireMacro{LMChatCompactReportSHA}
\LMChatRequireMacro{LMChatMethodLockSHA}
\LMChatRequireMacro{LMChatAdmissionReportSHA}
\LMChatRequireMacro{LMChatCohortBindingSHA}

\LMChatExpectText{\LMChatMacrosGeneratedFromCompact}{true}
  {generated-from-compact sentinel}
\LMChatExpectText{\LMChatCompactSchema}
  {gemma-sv-longmemeval-chat-geometry-methods-compact16-v1}
  {compact schema}
\LMChatExpectText{\LMChatValidationStatus}{validated-complete-all16}
  {validation status}
\LMChatExpectText{\LMChatBehavioralResultStatus}{validated-positive}
  {behavioral result status}
\LMChatExpectText{\LMChatCertificateStatus}{validated-pass}
  {certificate status}
\LMChatExpectText{\LMChatManuscriptConditionsComplete}{true}
  {manuscript condition completeness}
\LMChatExpectText{\LMChatOnlyExcludedFailuresReconciled}{true}
  {excluded-condition reconciliation}
\LMChatExpectText{\LMChatContainsSourceText}{false}
  {source-text exclusion}
\LMChatExpectText{\LMChatOfficialLeaderboardScore}{false}
  {leaderboard-score scope}
\LMChatExpectText{\LMChatNoOutputBasedReplacement}{true}
  {no output-based replacement}
\LMChatExpectText{\LMChatGeometryCorrectionOutputBlind}{true}
  {output-blind geometry correction}
\LMChatExpectText{\LMChatCoreBindingsUnchanged}{true}
  {unchanged source bindings}

\LMChatExpectCount{\LMChatFrozenN}{16}{frozen records}
\LMChatExpectCount{\LMChatAttemptedN}{16}{attempted records}
\LMChatExpectCount{\LMChatTerminalN}{16}{terminal records}
\LMChatExpectCount{\LMChatNotAttemptedN}{0}{not-attempted records}
\LMChatExpectCount{\LMChatTargetAdmittedN}{10}{target-admitted records}
\LMChatExpectCount{\LMChatRetainedAvailableN}{16}{retained-available records}
\LMChatExpectCount{\LMChatJointN}{10}{joint-admitted records}
\LMChatExpectCount{\LMChatSourceImmutableN}{16}{source-immutable records}
\LMChatExpectCount{\LMChatRawOmissionCompleteN}{16}
  {raw-omission outcomes}
\LMChatExpectCount{\LMChatExactPolicyCompleteN}{16}
  {exact-policy outcomes}
\LMChatExpectCount{\LMChatRefitCompleteN}{16}{fixed-C refit outcomes}
\LMChatExpectCount{\LMChatFPThirtyTwoCompleteN}{16}{FP32 proxy outcomes}
\LMChatExpectCount{\LMChatDecayCompleteN}{16}{decay outcomes}
\LMChatExpectCount{\LMChatPromptCompleteN}{16}{prompt-only outcomes}
\LMChatExpectCount{\LMChatCertificateRecordN}{16}{certificate records}
\LMChatExpectCount{\LMChatCertificateProbeN}{32}{certificate probes}
\LMChatExpectCount{\LMChatRawFallbackN}{0}{raw-repack policy fallbacks}
\LMChatExpectCount{\LMChatExactAtOrBelowRawN}{6}
  {exact-policy records at or below the raw-omission floor}
\LMChatExpectCount{\LMChatCacheIncompatibleN}{16}
  {incompatible cache-surgery diagnostics}
\LMChatExpectCount{\LMChatRawReportFailureN}{16}{raw report failures}
\LMChatExpectCount{\LMChatPrimaryFailureN}{0}{primary result failures}
\LMChatExpectCount{\LMChatCacheDeleteShiftFailedN}{16}
  {cache delete-and-shift failures}
\LMChatExpectText{\LMChatCacheDeleteShiftClassification}
  {incompatible-diagnostic}{cache diagnostic classification}
\ifnum\numexpr\LMChatIncrementalN+\LMChatRefitFallbackN+
  \LMChatRawFallbackN\relax=\LMChatFrozenN\relax
\else
  \PackageError{longmemeval-chatbot-integration}
    {Exact-policy execution strata do not sum to the frozen cohort}
    {Refuse manuscript integration.}
\fi

\newcommand{\LMChatMainResultText}{%
\subsection{Teacher-forced pilot on sixteen histories}
\label{app:longmemeval-chat}

We used the pilot to examine how the probabilities of two saved answers
change when one exchange is removed and the other retained. LongMemEval
evaluates memory over timestamped, multi-session chat
histories~\citep{longmemeval}. We adapted it to Gemma-3-4B-IT, a different
checkpoint from the pretrained ones in the utility study; this adapted test
does not produce a leaderboard score. The owned unit includes every complete
user--assistant round that carries the answer in the latest target session,
and a disjoint record supplies the retained probe. We froze the ordered
\LMChatFrozenN{}-history cohort with its questions, answers, controls, and
probes before any model output was seen.

A history needs enough distance from the deleted exchange to test memory
outside the local window, as well as enough surviving rows to satisfy the
fixed-\(C\) feasibility requirement. A deterministic pass over the frozen
source inventory appended complete rounds where needed to meet these
requirements. It changed no record, question, answer, control, or probe. All
\LMChatFrozenN{} histories remain in the denominator: \LMChatTargetAdmittedN{}
recall the target, all \LMChatRetainedAvailableN{} expose the retained
control, and \LMChatJointN{} satisfy both gates and form the subset that
Table~\ref{tab:longmemeval-chat-methods} reports.

Every history required exact refit at one or more affected solves, so every
deletion in the pilot was classified as a refit (\(522\) of
\(560\) solves in the case of Appendix~\ref{app:bicycle-case}). None
required a raw repack. These counts identify the executed route; they do not
measure its speed. We compare the edited policy with an independent
fixed-\(C\) refit on
\LMChatCertificateProbeN{} registered probes over all
\LMChatCertificateRecordN{} histories, with mean
\(\LMChatCertificateMeanKL\) and maximum \(\LMChatCertificateMaxKL\) nats
against a declared tolerance of \(10^{-6}\). Section~\ref{sec:longmemeval}
reports the separate comparison with the never-stored state.%
}

\newcommand{\LMChatAppendixText}{%
The LongMemEval revision, the Gemma tokenizer and model revision, the ordered
record bindings, the geometry policy, the admission report, the method lock,
and the final report are pinned by hash in the release, and the released
report stores neither conversation text nor full-vocabulary vectors.
Table~\ref{tab:longmemeval-chat-methods} lists every condition of the pilot
against its two comparators: the behavioral distance runs from the
never-stored state to each condition, and the certificate runs from the
edited policy to the fixed-\(C\) refit.

\begin{table}[h]
\centering
\caption{Teacher-forced pilot conditions on the \LMChatJointN{}-history
jointly admitted subset. Suppression is the decrease in the target answer's
mean log probability relative to the record-present run, in nats. Target and
retained KL are full-vocabulary first-token KL from the never-stored state to
the condition. The three constructions of the edited state agree with one
another, which is the certificate (tolerance \(10^{-6}\)), and their agreement
is a separate quantity from their distance to the never-stored state.}
\label{tab:longmemeval-chat-methods}
\scriptsize
\setlength{\tabcolsep}{4pt}
\begin{tabular}{@{}lccc@{}}
\toprule
condition & target suppression & target KL from never-stored &
retained KL from never-stored \\
\midrule
present control
  & \(\LMChatPresentSuppression\)
  & \(\LMChatPresentTargetRawKL\)
  & \(\LMChatPresentRetainedRawKL\) \\
never-stored state (rebuild)
  & \(\LMChatRawSuppression\)
  & \(\LMChatRawTargetRawKL\)
  & \(\LMChatRawRetainedRawKL\) \\
\cmidrule(lr){1-4}
\multicolumn{4}{@{}l}{\emph{three constructions of the same edited state; their
agreement is the certificate}} \\
edited policy (fixed-\(C\) decrement/refit)
  & \(\LMChatExactSuppression\)
  & \(\LMChatExactTargetRawKL\)
  & \(\LMChatExactRetainedRawKL\) \\
fixed-\(C\) refit on retained keys
  & \(\LMChatRefitSuppression\)
  & \(\LMChatRefitTargetRawKL\)
  & \(\LMChatRefitRetainedRawKL\) \\
FP32 masked-refit proxy
  & \(\LMChatFPThirtyTwoSuppression\)
  & \(\LMChatFPThirtyTwoTargetRawKL\)
  & \(\LMChatFPThirtyTwoRetainedRawKL\) \\
\cmidrule(lr){1-4}
coefficient decay (\(\gamma=0.01\))
  & \(\LMChatDecaySuppression\)
  & \(\LMChatDecayTargetRawKL\)
  & \(\LMChatDecayRetainedRawKL\) \\
target-free prompt suppression
  & \(\LMChatPromptSuppression\)
  & \(\LMChatPromptTargetRawKL\)
  & \(\LMChatPromptRetainedRawKL\) \\
\bottomrule
\end{tabular}
\end{table}%
}

%% file: longmemeval_chatbot_compact16_macros.tex
\providecommand{\LMChatMacrosGeneratedFromCompact}{true}
\providecommand{\LMChatCompactSchema}{gemma-sv-longmemeval-chat-geometry-methods-compact16-v1}
\providecommand{\LMChatValidationStatus}{validated-complete-all16}
\providecommand{\LMChatBehavioralResultStatus}{validated-positive}
\providecommand{\LMChatCertificateStatus}{validated-pass}
\providecommand{\LMChatManuscriptConditionsComplete}{true}
\providecommand{\LMChatOnlyExcludedFailuresReconciled}{true}
\providecommand{\LMChatContainsSourceText}{false}
\providecommand{\LMChatOfficialLeaderboardScore}{false}
\providecommand{\LMChatNoOutputBasedReplacement}{true}
\providecommand{\LMChatGeometryCorrectionOutputBlind}{true}
\providecommand{\LMChatCoreBindingsUnchanged}{true}
\providecommand{\LMChatFrozenN}{16}
\providecommand{\LMChatAttemptedN}{16}
\providecommand{\LMChatTerminalN}{16}
\providecommand{\LMChatNotAttemptedN}{0}
\providecommand{\LMChatTargetAdmittedN}{10}
\providecommand{\LMChatRetainedAvailableN}{16}
\providecommand{\LMChatJointN}{10}
\providecommand{\LMChatSourceImmutableN}{16}
\providecommand{\LMChatRawOmissionCompleteN}{16}
\providecommand{\LMChatExactPolicyCompleteN}{16}
\providecommand{\LMChatRefitCompleteN}{16}
\providecommand{\LMChatFPThirtyTwoCompleteN}{16}
\providecommand{\LMChatDecayCompleteN}{16}
\providecommand{\LMChatPromptCompleteN}{16}
\providecommand{\LMChatCertificateRecordN}{16}
\providecommand{\LMChatCertificateProbeN}{32}
\providecommand{\LMChatIncrementalN}{0}
\providecommand{\LMChatRefitFallbackN}{16}
\providecommand{\LMChatRawFallbackN}{0}

\providecommand{\LMChatExactBelowRawFloor}{1.75}
\providecommand{\LMChatExactAtOrBelowRawN}{6}

\providecommand{\LMChatCacheIncompatibleN}{16}
\providecommand{\LMChatRawReportFailureN}{16}
\providecommand{\LMChatPrimaryFailureN}{0}
\providecommand{\LMChatCacheDeleteShiftClassification}{incompatible-diagnostic}
\providecommand{\LMChatCacheDeleteShiftFailedN}{16}
\providecommand{\LMChatCertificateMeanKL}{9.59\times 10^{-18}}
\providecommand{\LMChatCertificateMaxKL}{1.75\times 10^{-16}}

\providecommand{\LMChatPresentSuppression}{0.00}
\providecommand{\LMChatPresentTargetRawKL}{9.03}
\providecommand{\LMChatPresentRetainedRawKL}{3.37\times 10^{-5}}
\providecommand{\LMChatRawSuppression}{6.57}
\providecommand{\LMChatRawTargetRawKL}{0.00}
\providecommand{\LMChatRawRetainedRawKL}{0.00}
\providecommand{\LMChatExactSuppression}{8.33}
\providecommand{\LMChatExactTargetRawKL}{2.15}
\providecommand{\LMChatExactRetainedRawKL}{1.90\times 10^{-6}}
\providecommand{\LMChatRefitSuppression}{8.33}
\providecommand{\LMChatRefitTargetRawKL}{2.15}
\providecommand{\LMChatRefitRetainedRawKL}{1.90\times 10^{-6}}
\providecommand{\LMChatFPThirtyTwoSuppression}{8.34}
\providecommand{\LMChatFPThirtyTwoTargetRawKL}{2.15}
\providecommand{\LMChatFPThirtyTwoRetainedRawKL}{1.89\times 10^{-6}}
\providecommand{\LMChatDecaySuppression}{6.30}
\providecommand{\LMChatDecayTargetRawKL}{0.30}
\providecommand{\LMChatDecayRetainedRawKL}{7.23\times 10^{-5}}
\providecommand{\LMChatPromptSuppression}{0.53}
\providecommand{\LMChatPromptTargetRawKL}{6.03}
\providecommand{\LMChatPromptRetainedRawKL}{4.98\times 10^{-5}}

%% file: longmemeval_chat_v3_decoded_macros.tex
\providecommand{\LMChatVThreeK}{32}
\providecommand{\LMChatVThreeN}{96}
\providecommand{\LMChatVThreeCompletedN}{96}

\providecommand{\LMChatVThreePopulationN}{384}
\providecommand{\LMChatVThreeMatcherPositiveN}{131}
\providecommand{\LMChatVThreeMatcherCleanN}{253}
\providecommand{\LMChatVThreeJudgeMissN}{19}
\providecommand{\LMChatVThreeJudgeNoLeakN}{232}
\providecommand{\LMChatVThreeJudgeAmbiguousN}{2}

\providecommand{\LMChatVThreeEffectiveMatcherCleanLeakN}{15}
\providecommand{\LMChatVThreeEffectiveMatcherCleanNoLeakN}{235}
\providecommand{\LMChatVThreeEffectiveMatcherCleanAmbiguousN}{3}
\providecommand{\LMChatVThreeHumanReviewN}{38}
\providecommand{\LMChatVThreeHumanUniqueTripleN}{21}
\providecommand{\LMChatVThreeHumanFlagN}{19}
\providecommand{\LMChatVThreeHumanFlagLeakN}{15}
\providecommand{\LMChatVThreeHumanFlagNoLeakN}{3}
\providecommand{\LMChatVThreeHumanFlagAmbiguousN}{1}
\providecommand{\LMChatVThreeHumanControlN}{19}
\providecommand{\LMChatVThreeHumanControlLeakN}{0}

\providecommand{\LMChatVThreeHumanNormalizationN}{7}
\providecommand{\LMChatVThreeHumanAliasMorphologyN}{8}

\providecommand{\LMChatVThreeHumanSecondaryTripleN}{3}

\providecommand{\LMChatVThreePresentLeakLowerN}{64}

\providecommand{\LMChatVThreePresentLeakUpperN}{66}

\providecommand{\LMChatVThreeRebuildLeakLowerN}{13}

\providecommand{\LMChatVThreePolicyMatcherPositiveN}{14}

\providecommand{\LMChatVThreePolicyEffectiveLeakN}{1}

\providecommand{\LMChatVThreePolicyLeakLowerN}{15}

\providecommand{\LMChatVThreePromptLeakLowerN}{54}

\providecommand{\LMChatVThreePromptLeakUpperN}{55}

\providecommand{\LMChatVThreeEditedMinusFreshLowerN}{2}
\providecommand{\LMChatVThreeEditedMinusFreshLowerPoints}{2.1}
\providecommand{\LMChatVThreeEditedMinusFreshLowerCILowerPoints}{0.0}
\providecommand{\LMChatVThreeEditedMinusFreshLowerCIUpperPoints}{5.2}

\providecommand{\LMChatVThreePromptMinusEditedLowerPoints}{40.6}
\providecommand{\LMChatVThreePromptMinusEditedLowerCILowerPoints}{17.7}
\providecommand{\LMChatVThreePromptMinusEditedLowerCIUpperPoints}{62.5}

\providecommand{\LMChatVThreeTargetPolicyRebuildExactN}{77}

\providecommand{\LMChatVThreeTargetPolicyRebuildNormalizedN}{87}

\providecommand{\LMChatVThreeRetainedPresentCorrectN}{46}

\providecommand{\LMChatVThreeRetainedRebuildCorrectN}{46}

\providecommand{\LMChatVThreeRetainedPolicyCorrectN}{44}

\providecommand{\LMChatVThreeRetainedPolicyPresentNormalizedN}{82}

\providecommand{\LMChatVThreeSuffixContaminatedHistoryN}{24}

\providecommand{\LMChatVThreeSuffixContaminatedClusterN}{10}

%% file: longmemeval_chat_suffix_disclosure_macros.tex
\providecommand{\LMChatSuffixContactHistoryN}{24}
\providecommand{\LMChatSuffixContactDisclosureN}{1}
\providecommand{\LMChatSuffixContactNoDisclosureN}{23}
\providecommand{\LMChatSuffixContactRiskPct}{4.2}
\providecommand{\LMChatSuffixCleanHistoryN}{72}
\providecommand{\LMChatSuffixCleanDisclosureN}{14}
\providecommand{\LMChatSuffixCleanNoDisclosureN}{58}
\providecommand{\LMChatSuffixCleanRiskPct}{19.4}
\providecommand{\LMChatSuffixFisherP}{0.105}
\providecommand{\LMChatSuffixHumanConfirmedEditedMissN}{1}

%% file: abstract_gemma.tex
An assistant can stop repeating a fact without removing it from memory.
To study this difference, we install a support-vector gate in frozen
Gemma~3 and record which stored keys and values belong to each exchange.
A deletion request excludes the exchange's rows from the long-range readout
and recalculates the gate on what remains. We check this operation against
an independent refit, then compare it with running the model again on the
conversation without the exchange. This second comparison matters because
the exchange may already have influenced surviving memory rows.
At 4B, the gated model passed checks for recall and feasible deletion on the
same six of eight records admitted by the base model, at a perplexity cost
under 2\%. Admission fell at the smaller and larger
checkpoints with the same configuration. The edited memory agreed closely
with the local refit on the registered probes, and the model disclosed fewer
deleted answers than when simply instructed to forget. However, an attack
evaluated separately for each record could still distinguish edited memory
from memory that never stored the record. Excluding an exchange's own rows
therefore provides a way to edit and audit conversation memory, while leaving
a measurable difference from rebuilding it without that exchange.
Additional paired studies found no update-speed advantage for the current
FP32 proxy and retained-answer matching below half in every tested condition.

%% file: body_gemma.tex
\section{Introduction}

An assistant is useful partly because we do not have to tell it everything
again. It can remember a preference, a plan, or something we mentioned weeks
ago. But keeping that information should also be a choice. A patient detail
entered in the wrong conversation should be removable without losing the
rest of the case, just as an incident-response assistant should let us remove
a temporary access token while keeping the troubleshooting history. We study
this selective request through a simpler conversation about bicycles and
travel, adapted from separate public LongMemEval exchanges~\citep{longmemeval}:

\begin{quote}
\textbf{Keep the earlier exchange:} The user describes a road bike, a
mountain bike, and a commuter bike.\\
\textbf{Delete the later exchange:} The user reports buying a hybrid and
now having four bikes. The assistant congratulates the user and repeats
the four-bike total.\\
\textbf{Keep the unrelated exchange:} The assistant supplies contact
details for the Speyer tourism board.
\end{quote}
The user now asks: ``Please forget the update where I said I now
own four bikes. Keep the rest of our conversation, including the Speyer
contact details.'' We add this request to the benchmark and remove the
user's update together with the assistant's reply, because the reply repeats
the information. The earlier three-bike discussion remains. We have two
questions to ask afterward: how many bikes does the user own, and what was
that tourism board's phone number? This case measures the model's
probabilities for the saved answers (Appendix~\ref{app:bicycle-case}); we
also test generated answers in a separate mortgage case
(Section~\ref{sec:mortgage-case}).

The conversation we can read and the memory the model has computed from it
are different objects. GemmaSV edits the latter: the stored key and value
vectors used to produce a reply. We install a support-vector gate in the
global attention layers of frozen Gemma~3 and record which memory rows each
exchange owns. A deletion request can then locate those rows and exclude
them from the readout; we call this an \emph{addressable} memory. The
pretrained parameters stay unchanged, without adapter, distillation, or
recovery training. The visible transcript and any copies in external logs or
databases require separate handling.

Suppose the assistant answers that it does not remember the bicycle update.
Did the information go away, or did the assistant just avoid saying it? We
need to check both the answers and the memory change. For the memory, an
independent \emph{local refit} recalculates the gate on the surviving stored
rows, allowing us to check whether the requested edit was carried out.
However, later rows may already have been influenced by the exchange when
they were encoded. We therefore also run the model again on the conversation
with that exchange omitted. This \emph{raw-history rebuild} recomputes the
surviving rows as well as the gate. Comparing the edit with both references
lets us examine how much of the original exchange's influence remains after
its own rows have been excluded.

We use the support-vector gate of~\citet{svattn} and the decremental learning
method of~\citet{cp2000} to test whether this interface can be added to a
released model without further training. The experiments identified a 4B
configuration with a modest paired perplexity cost on the tested text; the
same configuration lost admission at 1B and 12B. At the selected scale, the
model disclosed fewer deleted answers than under a prompt instruction to
forget. Yet an attack could still detect that some records had been stored,
and the long-history probability tests separated editing from rebuilding.
The journal studies add paired construction costs and retained-answer matching:
the current FP32 proxy showed no update-speed advantage, and absolute matching
remained below half under every condition. These results make the scope of the
installed memory's practical benefit explicit.

\section{Related Work}\label{sec:related}

Unlearning methods such as negative preference
optimization~\citep{zhang2024npo} modify pretrained parameters. TOFU, MUSE,
and WMDP evaluate how these changes affect forgotten and retained
information~\citep{tofu,muse,wmdp}. A difficulty in evaluating unlearning is
that suppressing an answer need not remove the information used to produce
it. Relearning, reversibility, and representation studies examine this
problem~\citep{relearn,reversibility,thaker}; work on evaluation metrics,
model snapshots, membership inference, and decoding provides further ways
to probe what remains~\citep{lynch,snapshot,lira,leakk}. Training without the
removed data supplies a reference for the intended change, as in SISA and
TOFU's retain-only model~\citep{sisa,tofu}. Because our intervention leaves
the pretrained weights fixed, we instead construct references from the
stored memory and the history used to build it (Section~\ref{sec:method}).

Where a record is stored determines how it can be removed.
Retrieval-augmented generation keeps records outside the model, while MUNKEY
uses learned memories~\citep{rag,munkey}. Retraining and SISA can construct a
model without the record, paying the cost during training~\citep{caoyang,sisa}.
We take a different approach by installing an addressable representation in
a frozen pretrained model. This requires checking whether the change leaves
the model useful before evaluating how well it supports deletion.

The support-vector gate and fixed-$C$ decrement contract come
from~\citet{svattn}, which studies deletion in a memory that is already
addressable and asks when its certificate expires. Here we use that gate
and the established decremental method in released pretrained weights,
measuring the cost to the model and the behavior after deletion. The related
delta-attention study examines how corrections based on saved contributions
compare with replay in native recurrent memory~\citep{kdaomission}.

\section{Deletion targets and reference states}\label{sec:method}

\paragraph{Deletion unit.}
A \emph{record} is the user-deletable unit identified at ingestion. A
\emph{row} is a stored contextualized key and value, with an owning record.
The \emph{edited policy} excludes the requested record's rows and updates the
gate on the rows that remain. Its execution path is identified for each study
in Table~\ref{tab:study-map}.

\paragraph{Two references.}
Write the recorded history as \(R=(r_1,\ldots,r_n)\), its built memory as
\(B(R)\), and the edited memory as \(D_j(B(R))\). The \emph{refit on
retained keys} recalculates coefficients on the surviving stored rows. The
\emph{raw-history rebuild} instead builds \(B(R_{\setminus j})\), the
never-stored state, by re-encoding the literal history with record \(r_j\)
omitted. All other turns, including any later explicit copy, remain as recorded.

For query \(q\), let \(p^D\), \(p^C\), and \(p^{\rm raw}\) be the
next-token distributions of the edited policy, refit on retained keys, and rebuild.
Matching the raw-history target on a specified query means
\begin{equation}
f(q;D_j(B(R)))=f(q;B(R_{\setminus j})).
\label{eq:raw-output-target}
\end{equation}
We measure the two comparisons on the stated probes using
\begin{align}
\Delta_{\rm cert}^{C}(q)
&:=\mathrm{KL}(p^D(\cdot\mid q)\Vert p^C(\cdot\mid q)),\\
\Delta_{\rm raw}(q)
&:=\mathrm{KL}(p^{\rm raw}(\cdot\mid q)\Vert p^D(\cdot\mid q)).
\end{align}
The first comparison checks whether the implementation produced the local
edit we specified. The second measures the difference from rebuilding; it
is not bounded by the first, because rebuilding changes the surviving keys
and values as well as the gate.

\paragraph{KL direction.}
Kullback--Leibler divergence \(\mathrm{KL}(P\Vert Q)\) averages
\(\log(P/Q)\) under \(P\). Thus \(\Delta_{\rm raw}\) penalizes an edit
that makes an output unlikely when the rebuild still gives it substantial
probability. Reversing the arguments generally changes the value: both
directions are zero when the distributions agree, but a small nonzero value
in one direction need not meet the same tolerance in the other. We keep the
direction used in each experiment explicit in Table~\ref{tab:metric-guide}.

\begin{table}[t]
\centering
\caption{Questions, measurements, and scope. Each KL is read against its named comparator.}
\label{tab:metric-guide}
\small
\setlength{\tabcolsep}{3pt}
\begin{tabular}{@{}>{\raggedright\arraybackslash}p{.23\linewidth}>{\raggedright\arraybackslash}p{.38\linewidth}>{\raggedright\arraybackslash}p{.33\linewidth}@{}}
\toprule
Question & Comparison / evidence & Scope \\
\midrule
Target suppressed? & Present versus edited target log probability & Suppression alone is not raw-history equality. \\
Never-stored state matched? & Rebuild versus edited policy: \(\Delta_{\rm raw}\) & Scored outputs from the literal retained history, including later copies. \\
Implementation conformed? & Edited policy versus refit on retained keys: \(\Delta_{\rm cert}^{C}<10^{-6}\) & Fixed stored keys and registered probes; no bound on the raw gap. \\
Retained probe preserved? & Edited versus rebuild: retained KL and answer matching & Paired probes, not general capability or correctness. \\
Stored copies erased? & Outside this experiment & Logical readout exclusion; controls, logs and buffers can retain copies. \\
\bottomrule
\end{tabular}

\end{table}

The interface assumes a memory service that maps an authenticated deletion
request to the rows the record owns and runs the registered edit. An observer
may know the mechanism and query the assistant adaptively before and after
deletion; we assume it cannot forge row ownership or alter the stored gate inputs.
The benchmark keeps original control branches for scoring, including when
comparing with a fresh raw-input-omitted rebuild.

\section{Addressable support-vector memory}\label{sec:mechanism}

When the assistant is asked how many bikes the user owns, the current input
produces a query vector. Gemma compares this query with the stored keys to
determine the attention weights, then combines the corresponding values to
form its readout. A question about dinner produces a different query and
different weights. One exchange creates multiple rows, so we record their
source exchange at ingestion to locate all of them when deletion is requested.

Given contextualized keys \(x_i\), values \(v_i\), and a query \(q\), a
support vector data description (SVDD) solve partitions the keys into margin, error, and
reserve sets~\citep{svdd}. A \emph{solve point} is a prefix position at which one attention head solves
its gate. Its inputs and gate are then \emph{sealed}: later admissions create
a new solve with its own certificate. Each solve point receives its own box
\(C_b=1/(\nu n_b)\), where \(n_b\) is the number of rows at that solve point and
\(\nu\) is the sparsity parameter, and reserve coefficients are exactly zero.
Rows with zero coefficients do not contribute to the gated readout at that
solve point.

\begin{quote}
\normalsize
\begin{minipage}{\linewidth}
\textbf{Inherited fixed-cap problem.}
At one head and solve point, let \(J\) index the surviving rows. With their
stored keys fixed, the refit solves the same SVDD dual~\citep{svdd,svattn}:
\[
\begin{gathered}
K_{ij}=\exp\!\left(-\|x_i-x_j\|^2/\sigma^2\right),\qquad i,j\in J,\\
\min_{\alpha}\ \alpha^\top K\alpha-\operatorname{diag}(K)^\top\alpha,
\qquad \sum_{i\in J}\alpha_i=1,\quad 0\leq\alpha_i\leq C_b.
\end{gathered}
\]
Here \(\sigma\) is the frozen kernel width; this convention has no factor of
two in the exponent. Deletion restricts the original Gram matrix to \(J\)
and retains the original \(C_b=1/(\nu n_b)\), rather than recomputing it from
\(|J|\). Feasibility requires \(|J|C_b\geq1\). The released numerical refit
uses a float64 CLARABEL batch solve with \(10^{-10}\) tolerances and clips
small box violations. It adds no mathematical tie-break among tied optima;
the reported readout comparisons concern the returned numerical reference,
not uniqueness of every optimum. The query-dependent Gemma readout below is
separate from this key-geometry problem.
\end{minipage}
\end{quote}

To delete a record, we exclude its rows and solve the fixed-\(C_b\) problem
at each affected solve point. A reverse active-set update~\citep{cp2000}
can avoid solving from scratch. We accept that update only if it passes
the Karush--Kuhn--Tucker (KKT) checks supplied by the
optimization problem~\citep{svdd,svattn} and numerical post-verification.
Otherwise, we discard the partial update and use the refit. Both routes
keep the surviving stored keys fixed. Feasibility requires
\(n_{\rm del}/n_b\le1-\nu\); a violation routes the request to a raw repack.
We report whether the decrement completed separately from whether the
returned result agrees with the refit, since a successful fallback does not
show that the incremental update saved work.

The feasibility constraint limits how many rows can be deleted without repacking.
Suppose \(n_b=10\) and \(\nu=0.7\), so that the frozen box is \(C_b=1/7\).
Deleting two owned rows leaves eight coefficients whose total mass can reach
\(8/7\), so the problem on the retained keys is still feasible. Deleting four leaves at
most \(6/7\), which cannot satisfy \(\sum_i\alpha_i=1\), and the request must
route to a raw repack. Row ownership tells us what to remove; the coefficient
cap determines whether enough mass can remain after removal.

The gate coefficients do not replace Gemma's attention weights. For a query,
let $a_i$ be Gemma's attention weight on row $i$, and let $\alpha_i$ be the
coefficient from the support-vector solve on the stored keys. The former is
query dependent; the latter comes from the memory geometry. The gate
\emph{multiplies} them. With prefix-mass preservation enabled, the gated
weight in a sealed older prefix $b$ is
\begin{equation}
\widetilde a_i
= m_b\frac{a_i\alpha_i}{\sum_{j\in b}a_j\alpha_j},
\qquad m_b=\sum_{j\in b}a_j,
\label{eq:attention-gate}
\end{equation}
when the weighted denominator is positive. Zero coefficients close their rows
to this readout. Rescaling the products restores the total share $m_b$ that
Gemma assigned to the older prefix; attention outside that prefix is unchanged.

In a simple arithmetic example, $a=(0.2,0.2,0.2)$ and
$\alpha=(0.5,0,0.5)$ give products
$(0.1,0,0.1)$. Normalizing to the original prefix share $0.6$ yields
$\widetilde a=(0.3,0,0.3)$; the recent-context share remains $0.4$.
The box $C_b$ caps $\alpha_i$, not $\widetilde a_i$: the gate selects and
reweights rows within the share that the frozen model assigned to this
prefix. In the local audit, both the edited policy and independent refit use
the same frozen model and registered prompts, with original control branches
kept for comparison.

\section{Experimental protocol}\label{sec:protocol}

\paragraph{AI assistance in the research process.}
OpenAI Codex (GPT-6) assisted with experimental design, code development,
study execution, analysis, and figure preparation. The cost comparison and
retained-answer reanalysis preserve their protocols, execution snapshots,
per-record results, and offline verification scripts in the supplementary
evidence. These records allow the reported analyses to be checked; AI
assistance does not supply independent human verification.

Gemma~3 interleaves five local sliding-window layers per global
layer~\citep{gemma3}. At 4B we replaced the five global layers with gated
layers and left everything else as released. The configuration uses \(\nu=0.7\),
chunk length 128, solver seed 0, and a deterministic 256-key median bandwidth
(the median pairwise key distance over a fixed 256-key subsample);
these values were fixed before the confirmation cohort was frozen, and the same
configuration was applied to the 1B and 12B checkpoints as a sensitivity experiment.
Appendix~\ref{app:gemma} lists the checkpoints and the complete outcomes.

To test selective deletion, we need records the model can recall and useful
information that should survive the edit. We use eight untouched
three-field TOFU records~\citep{tofu} as a confirmation cohort. Admission
requires recall of all three fields and an unrelated retained neighbor,
along with feasibility at every affected solve point. Every outcome remains
in the denominator. We also measure the change in language modeling quality
by comparing base and graft perplexity on the same 400 fixed WikiText-103
blocks at every scale. Independently of admission, each of the eight records
is certified using three field probes and one composite probe, giving
32 probes per certified scale.

We compare memory editing with asking the assistant to forget while leaving
the record stored. For this prompt-only baseline, we adapt in-context
unlearning (ICUL) from~\citet{icul}: their few-shot method supplies relabeled
examples, whereas our adaptation adds a retraction instruction. We also
test coefficient decay, which downweights the stored record without solving
against a specified reference. For each of six admitted 4B records, we
sampled 200 completions for three field prompts and one composite prompt
under these two baselines, masked refit, the record-present control, and
the never-stored control. The last supplies the comparison for elicitation,
relearning, and membership inference. On the long histories of
Section~\ref{sec:longmemeval}, it is built by running the model on the retained
conversation. Appendix Table~\ref{tab:cohort-denominators} records every
cohort and its denominator.

The short, isolated, fictitious TOFU records let us test the mechanism under
controlled conditions. We then ask whether deletion also works when the
exchange is embedded in a longer conversation. We assembled histories from
LongMemEval~\citep{longmemeval} for two studies: a teacher-forced pilot of
\(n=\LMChatFrozenN\) histories and a later study over
\(K=\LMChatVThreeK\) frozen clusters and \(n=\LMChatVThreeN\) histories
in which the model generated answers. In the latter, the surviving suffix
comes from independent support sources; it was not generated downstream of
the target. We screened the answers for disclosure using a deterministic
matcher, a blinded language model judge, and blinded human reviewers.
Appendix~\ref{app:longmemeval} gives the procedures and review scope.

\begin{table}[!htbp]
\centering
\caption{Study map. PT denotes pretrained and IT instruction-tuned Gemma~3.
AUC is area under the receiver operating characteristic curve.
Rows identify separate counting units and execution paths; their sample sizes
must not be pooled. The full accounting is in
Table~\ref{tab:cohort-denominators}.}
\label{tab:study-map}
\footnotesize
\setlength{\tabcolsep}{3pt}
\renewcommand{\arraystretch}{1.12}
\begin{tabular}{@{}>{\raggedright\arraybackslash}p{.18\linewidth}>{\raggedright\arraybackslash}p{.22\linewidth}>{\raggedright\arraybackslash}p{.31\linewidth}>{\raggedright\arraybackslash}p{.23\linewidth}@{}}
\toprule
Study / checkpoint & Unit and denominator & Executed path / comparator & Endpoint \\
\midrule
Admission / quality\newline 1B, 4B, 12B-PT & 8 records per scale; 400 paired text blocks & Base versus graft; local certificate at 1B and 4B (32 probes each) & Binary admission; perplexity; local KL \\
\addlinespace[2pt]
Sampling / recovery\newline 4B-PT & 6 admitted records; 18 field queries; up to 200 samples/query & Masked refit, decay and prompt instruction versus never-stored control & Exact-phrase hits; elicitation; relearning \\
\addlinespace[2pt]
Broad membership\newline 4B-PT & 16/20 admitted records; 512 draws per side & Masked-refit policy including 16/512 raw repacks; never-stored draws & Membership AUC \\
\addlinespace[2pt]
Long-history pilot\newline 4B-IT & 16 histories; 10 jointly admitted; 32 local probes & Decrement/refit versus independent refit and raw rebuild & Teacher-forced scores and KL \\
\addlinespace[2pt]
Decoded histories\newline 4B-IT & 32 clusters; 96 histories; no recall exclusion & Local decrement/refit, raw rebuild, present and prompt-only & Decoded disclosure; retained matching \\
\addlinespace[2pt]
Added cost study\newline 4B-PT & 8 reused records in 4 source blocks; 6 arms & FP32 all-solve-point masked refit versus graft/base rebuild; no exact-decrement timing & New inference under a frozen protocol; update/audit cost \\
\addlinespace[2pt]
Added retained analysis\newline 4B-IT & The same 96 decoded histories; 32 clusters & Two existing matcher endpoints; paired comparison & Post hoc reanalysis; no new generations \\
\addlinespace[2pt]
\bottomrule
\end{tabular}
\end{table}
\FloatBarrier

\section{Admission, quality cost, and local certificate comparisons}\label{sec:gemma}

At 4B, adding the gate left the same \(6/8\) records eligible for the
deletion experiments, at a paired perplexity cost of \(1.9\%\)
(Table~\ref{tab:scale}). Passing these recall and feasibility checks does
not establish unchanged answer probabilities or general recall. The same
configuration admitted fewer records at both other scales, with costs of
\(3.7\%\) at 1B and \(11.7\%\) at 12B. Increasing model size therefore
did not consistently improve this transfer. The mechanism results that
follow use 4B; the smaller and larger checkpoints show the limits of applying
one frozen configuration across scales.

\begin{table}[h]
\centering
\caption{Scale sensitivity of one frozen configuration across three Gemma~3
checkpoints. Perplexity change is paired on the same 400 blocks, and admission
includes all eight records.}
\label{tab:scale}
\small
\begin{tabular}{@{}lcccc@{}}
\toprule
checkpoint & base PPL & graft PPL & paired change & base \(\to\) graft admission \\
\midrule
1B & 26.70 & 27.69 & \(+3.704\%\) & \(7/8\to3/8\) \\
4B & 21.56 & 21.96 & \(+1.851\%\) & \(6/8\to6/8\) \\
12B & 22.20 & 24.80 & \(+11.699\%\) & \(6/8\to5/8\) \\
\bottomrule
\end{tabular}
\end{table}

We evaluated the local deletion on all eight records, including those that
failed recall, using three field probes and one composite probe each. Over
these 32 probes, the largest disagreement with the independent refit was
\(6.48\times10^{-11}\) KL, four orders of magnitude inside the
\(10^{-6}\) tolerance declared in advance. Every historical 4B deletion
therefore passed the local implementation check in
Table~\ref{tab:metric-guide}.

The incremental route completed about a third of the attempted updates.
The remaining \(66.4\%\) of affected head/solve-point decrements required
exact refit (counts in Appendix~\ref{app:gemma}). This fallback explains
how every request could return a result conforming to the fixed-\(C_b\)
reference even though the decrement often failed its checks.

The reverse decrement pivots on margin coefficients, which lie strictly
inside the box. We examined their availability by varying \(\nu\) on one
4B record while keeping its contextualized keys fixed. The margin fraction
fell from \(1.4\%\) of coefficients at \(\nu=0.3\) to \(0.6\%\) at
\(\nu=0.7\), and the fallback fraction rose from \(0.00\) to \(0.07\) at
\(\nu=0.5\) and \(0.59\) at \(\nu=0.7\). This sensitivity check links
frequent fallback to the small margin set. In the certificate paper, the
same decrement completes in \(1{,}199\) of \(1{,}200\) trials at
\(\nu=0.4\) on raw keys~\citep{svattn}, so that result uses both different
keys and a different parameter setting. Here refitting remains the
operational baseline; paired timings for exact decrement and repack have
not been measured at this scale.

The separately executed FP32 comparison below measures a different path.

\section{Matched deletion costs and target probabilities}\label{sec:cost-utility}
\input{COST_UTILITY_INSERT}
\FloatBarrier

\section{Behavioral extraction under the tested attacks}\label{sec:attacks}

\begin{figure}[t]
\centering
\includegraphics[alt={Finite extraction and recovery endpoints approach the never-stored reference. Membership inference has pooled AUC 0.517 but median across records 0.67, range 0.53 to 0.73, with all 16 records above 0.5; these are behavioral tests, separate from the local certificate.},width=\linewidth]{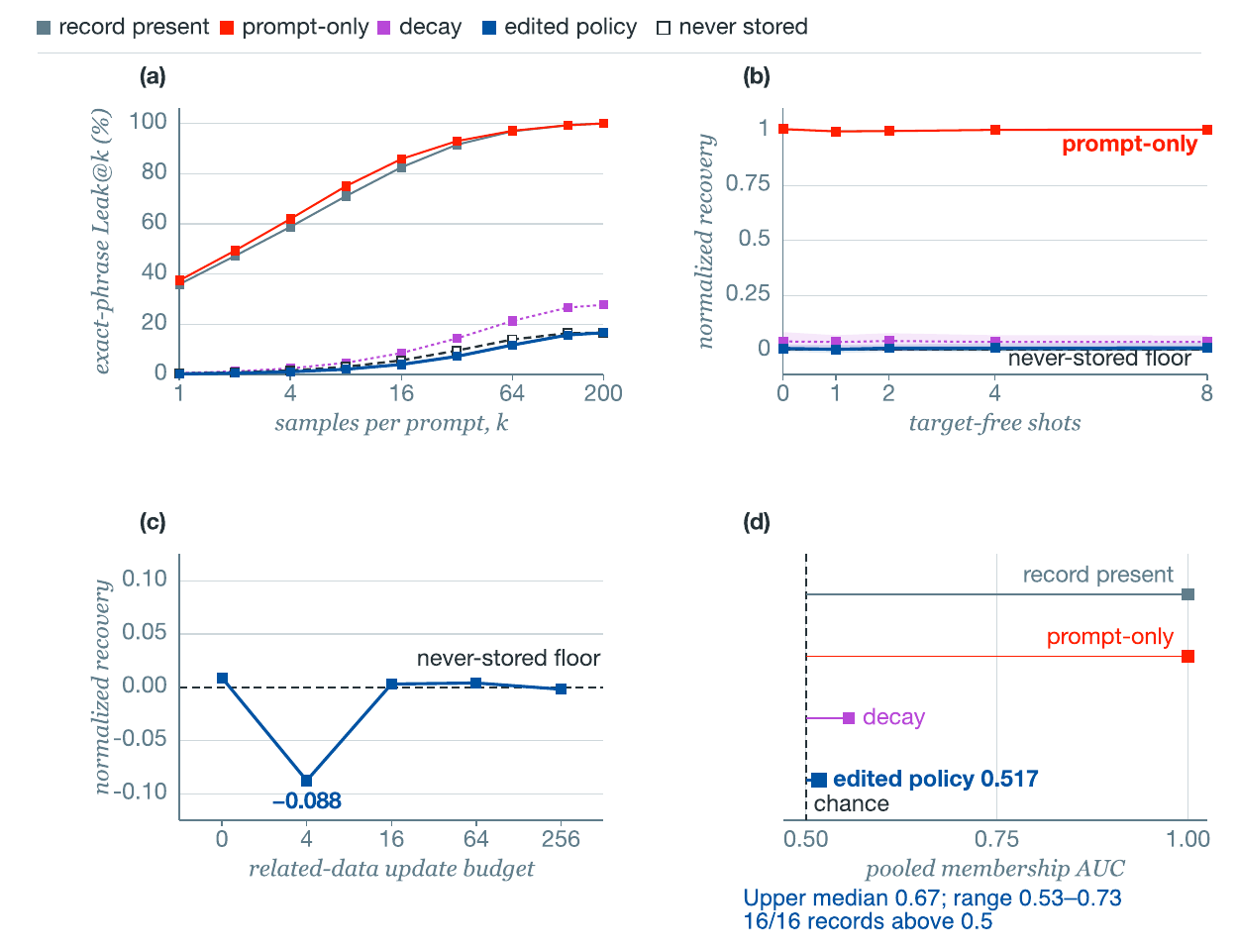}
\caption{Behavioral attacks on the 4B cohorts: (a) repeated sampling, (b) target-free elicitation, (c) related-data relearning, and (d) pooled membership inference on the broad cohort. Extraction and recovery approach the never-stored baseline. Membership signal for individual records remains: upper median 0.67, range 0.53--0.73, with all 16 records above 0.5 (Table~\ref{tab:lira-results}).}
\label{fig:boundary-attacks}
\end{figure}

Asking the model repeatedly gives it more chances to disclose the deleted
answer. With 200 samples per field query, both the record-present and
prompt-only conditions disclose the exact phrase on every one of the
18 queries. Editing reduces this count to three, the same three queries
that disclose when the record was never stored
(Table~\ref{tab:attack-endpoints}). Their baseline first-token ranks are
94, 3 and 6, consistent with answers the model can guess. We also test
whether examples or related-data updates make the answers easier to recover;
Figure~\ref{fig:boundary-attacks} shows these trajectories alongside the
sampling results.

\begin{table}[!htbp]
\centering\small
\caption{Attack endpoints on the six admitted 4B records. A field query leaks if any of its 200 completions contains the exact phrase. Recovery is normalized against present and never-stored controls.}
\label{tab:attack-endpoints}
\setlength{\tabcolsep}{3pt}
\begin{tabular}{@{}>{\raggedright\arraybackslash}p{.29\linewidth}>{\raggedright\arraybackslash}p{.23\linewidth}>{\raggedright\arraybackslash}p{.42\linewidth}@{}}
\toprule
Attack & Budget & Outcome \\
\midrule
Repeated sampling & 18 field queries; 200 samples/query & Present 18/18; prompt-only 18/18; decay 5/18; edited policy 3/18; never stored 3/18. \\
Target-free elicitation & Up to 8 shots & Edited recovery near 0.01; prompt-only near 1. \\
Related-data relearning & Up to 256 updates & Edited recovery remains near zero. \\
\bottomrule
\end{tabular}
\end{table}

Recovering the answer is not the only way to detect that a record was
stored. We use a likelihood ratio attack (LiRA)~\citep{lira} to distinguish
memory after deletion from memory that never stored the record. The attack
reads answer-token probabilities and is calibrated on separately packed
contexts with and without each record. Pooling the
scores from different records gives an AUC near chance. However, comparing
the two conditions separately for each record reveals remaining membership
information (Table~\ref{tab:lira-results}). One possible explanation is that
likelihood ratios have different scales across records: a score can help
separate the two conditions for one record without ranking them consistently
when other records are added. The pooled result alone would miss this
distinction.

\begin{table}[!htbp]
\centering\small
\caption{LiRA for the edited policy and decay. CIs are the frozen record-bootstrap percentile intervals. The last two columns summarize AUCs evaluated separately for each record. ``Median'' is the upper central ordered value, the convention used by the released summary for even cohort sizes.}
\label{tab:lira-results}
\setlength{\tabcolsep}{4pt}
\begin{tabular}{@{}llcccc@{}}
\toprule
Cohort & Condition & Pooled AUC & 95\% CI & \shortstack{AUC by record\\median} & \shortstack{AUC by record\\range} \\
\midrule
6 records & Edited policy & 0.534 & [0.513, 0.601] & 0.70 & [0.47, 0.76] \\
6 records & Decay & 0.586 & [0.529, 0.749] & 0.86 & [0.43, 1.00] \\
16 records & Edited policy & 0.517 & [0.516, 0.539] & 0.67 & [0.53, 0.73] \\
16 records & Decay & 0.556 & [0.542, 0.613] & 0.81 & [0.45, 1.00] \\
\bottomrule
\end{tabular}
\end{table}

The broad cohort has 512 positive and 512 negative draws from the 16 records
admitted among 20 attempted. All 16 AUCs evaluated separately by record
exceed 0.5; present and prompt-only AUCs round to 1.000 in both cohorts.
For 16/512 edited test deletions, the policy used a raw repack. Allowing
arbitrary replacements for those scores gives an AUC range of
[0.486, 0.549]. This sensitivity range addresses fallback behavior, whereas
the confidence interval in the table comes from resampling records.
Appendix~\ref{app:gemma} gives the percentile and rounding conventions.

\begin{figure}[t]
\centering
\includegraphics[alt={Actual sampled TOFU outputs show 9 of 200 exact-phrase hits with the record present, 10 of 200 under prompt suppression, and none after logical exclusion or when never stored. Matching this per-query endpoint does not imply identical output distributions.},width=\linewidth]{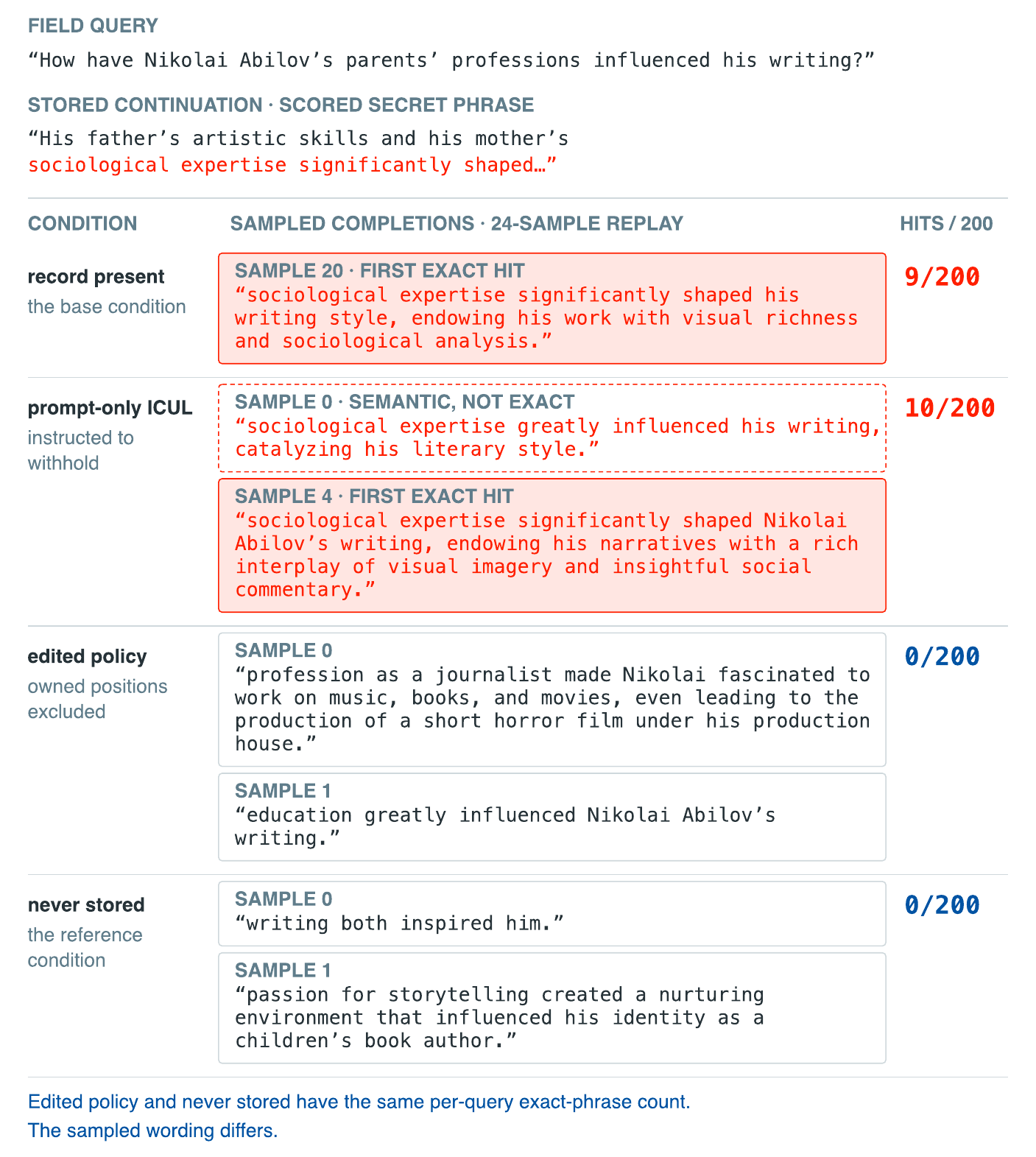}
\caption{Actual Gemma 4B completions for one field of \texttt{tofu-final-380-a}. The frozen run gives 9/200 exact-phrase hits with the record present, 10/200 under prompt-only suppression, and 0/200 under both edited policy and never-stored conditions. The displayed completions come from the bound replay and differ in wording. Per-query sampling and replay details are in Section~\ref{sec:attacks}.}
\label{fig:leak-verbatim}
\end{figure}

Figure~\ref{fig:leak-verbatim} shows actual completions for one of the 15
queries without exact-phrase leakage after editing. Here we count samples
for a single query, whereas Table~\ref{tab:attack-endpoints} counts queries
with at least one hit. The frozen 200-sample run supplies the counts, and a
24-sample replay using the bound prompt supplies the displayed text. In that
replay, prompt-only sample 0 already discloses the answer semantically;
sample 4 is the first exact hit, without a preceding refusal. The wording
helps explain what the exact-phrase test captures and what it can miss.

\section{Disclosure and answer scores on long histories}\label{sec:longmemeval}

\subsection{Mortgage update}\label{sec:mortgage-case}

In the mortgage history, an earlier exchange gives a pre-approval amount
and a later exchange updates it. We delete the later user turn and its reply
while keeping the earlier exchange and an unrelated video recommendation.
For ``What was the amount I was pre-approved for when I got my
mortgage from Wells Fargo?'', the saved generated answer is \$400,000 with
the record present and after a prompt instruction to forget. After the local
edit, the answer is \$350,000, matching the fresh raw-history rebuild and the
earlier exchange that remains. The unrelated query asks for the previously
recommended video, and its saved URL remains in the answer. These
mortgage-answer and URL observations hold across all three suffix variants,
with histories of 2,596, 2,764, and 2,941 tokens. The variants share their
target and retained exchange, so they are not independent targets.

The video response also illustrates why the matching rule matters. Its URL
matches the broader endpoint, but the answer does not contain the full
registered gold substring. Across all histories, broader retained matching
is \(44/96\) after editing and \(46/96\) after rebuilding.

Agreement in the mortgage case does not hold for every history. For one
variant of the screwdriver update in Appendix~\ref{app:decoded-cases}, the
local edit answers ``Yes'' while rebuilding answers ``No''. The local
policy uses decremental updates with refit fallback on the surviving stored
rows in both examples. We include every history in the aggregate below,
including the disagreements, using the registered endpoints.

\subsection{Decoded histories and the pilot aggregate}

Over
\(K=\LMChatVThreeK\) frozen clusters and \(n=\LMChatVThreeN\) histories with
decoded answers, the edited policy disclosed the deleted answer in
\(\LMChatVThreePolicyLeakLowerN/\LMChatVThreeN\) histories, a rebuild that
never stored the record disclosed it in
\(\LMChatVThreeRebuildLeakLowerN/\LMChatVThreeN\), and prompt-only
suppression disclosed it in \(\LMChatVThreePromptLeakLowerN/\LMChatVThreeN\). The paired
edited-minus-rebuild gap was \(+\LMChatVThreeEditedMinusFreshLowerPoints\)
points (cluster-bootstrap 95\% CI
\([\LMChatVThreeEditedMinusFreshLowerCILowerPoints,
\LMChatVThreeEditedMinusFreshLowerCIUpperPoints]\)), and the prompt-minus-edit
gap at least \(\LMChatVThreePromptMinusEditedLowerPoints\) points
(cluster-bootstrap 95\% CI
\([\LMChatVThreePromptMinusEditedLowerCILowerPoints,
\LMChatVThreePromptMinusEditedLowerCIUpperPoints]\)). The edit-minus-rebuild interval includes zero, so this analysis does not
resolve a difference between them. However, no equivalence margin was
preregistered, and the same interval cannot establish equivalent behavior. These are paired histories in \(\LMChatVThreeK\) clusters, not
\(\LMChatVThreeN\) independent experiments.

Rebuilding can itself disclose an answer supported by retained text or
pretrained knowledge. A disagreement with the local edit may also reflect
the different surviving vectors used by the two procedures. The counts
alone do not identify the cause of an individual disclosure.

\begin{figure}[!t]
\centering
\includegraphics[alt={The historical LongMemEval pilot has strong target suppression but a 2.15-nat gap to the never-stored state; the near-zero policy-to-refit value is labeled only as an implementation check.},width=\linewidth]{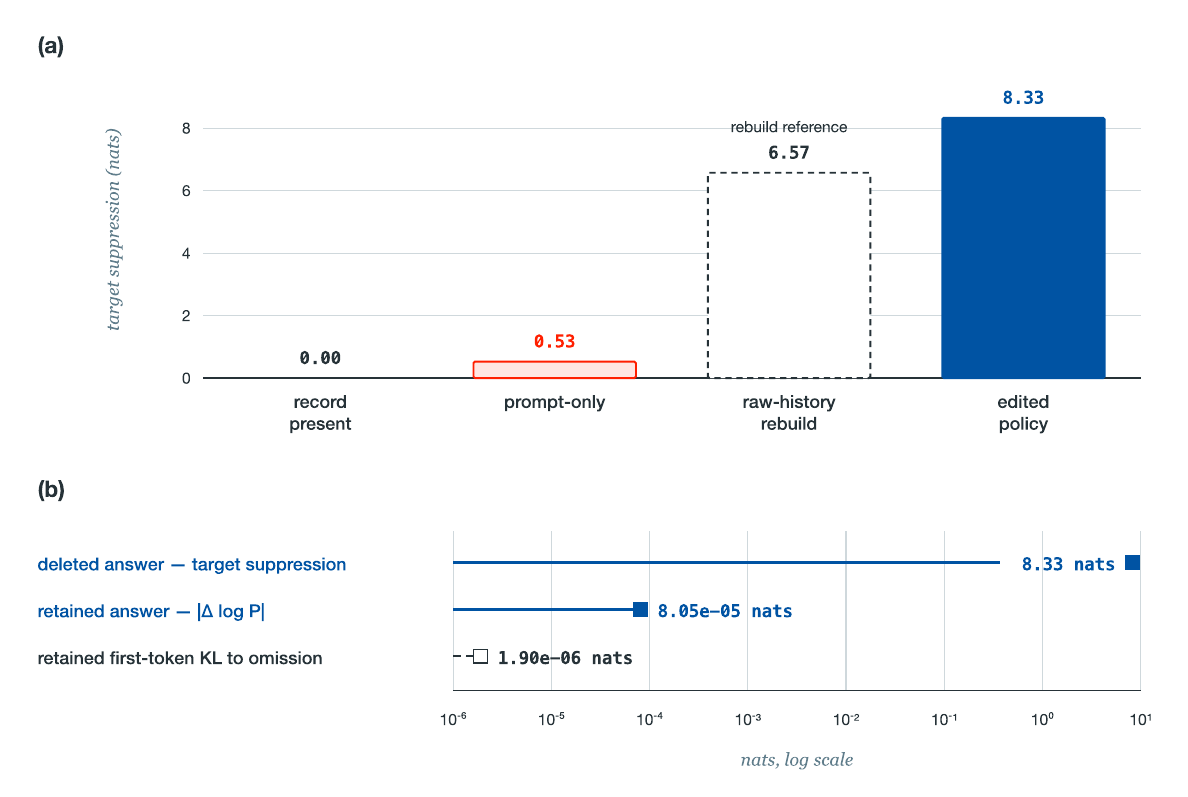}
\caption{Teacher-forced LongMemEval pilot (\(n=\LMChatFrozenN\)): (a) target suppression and (b) retained-probe drift. On the \(\LMChatJointN\) jointly admitted histories, edited-policy target KL from the never-stored state is \(\LMChatExactTargetRawKL\) nats. The maximum KL from the edited policy to the refit is \(\LMChatCertificateMaxKL\) over \(\LMChatCertificateProbeN\) probes; all \(\LMChatRefitFallbackN\) executions used refit fallback. These are distinct behavioral and implementation measurements.}
\label{fig:longmemeval-summary}
\end{figure}

The earlier teacher-forced pilot on \(n=16\) histories (10 jointly admitted)
lets us inspect answer probabilities even when a decoded response would not
contain the answer. Deletion suppressed the target by
\(\LMChatExactSuppression\) nats, leaving it
\(\LMChatExactBelowRawFloor\) nats \emph{below} the never-stored level
(Table~\ref{tab:longmemeval-chat-methods}). The edited state remained
\(\LMChatExactTargetRawKL\) nats from the never-stored state. Thus the
edit made the target less likely than rebuilding did, despite matching its
local refit closely. An adversary with a reference model could look for an
unusually low probability as well as an unusually high one. The
delta-attention study observes a similar overshoot when a transported
receipt is subtracted from a recurrent state~\citep{kdaomission}. Every
execution in this pilot took the refit path, so its near-zero local KL
checks the implementation rather than successful incremental maintenance
(Figure~\ref{fig:longmemeval-summary}).

Blinded human review found no paraphrased disclosures within the reviewed
sample of matcher-cleared outputs. Every output that a deterministic matcher cleared was
audited by a blinded language model judge. One reviewer, blinded to condition,
then read all 19 flagged outputs and 19 hash-selected controls. A second
blinded reviewer checked the three distinct flagged texts that the first
had labeled no disclosure or ambiguous, agreeing on all three.
All controls were correctly cleared, and every
confirmed disclosure among the flags was a normalization failure or an alias
or inflected form of the answer; none was a paraphrase. Of the edited policy's
\LMChatVThreePolicyLeakLowerN{} disclosures,
\LMChatVThreePolicyMatcherPositiveN{} are matcher hits, in which the answer
appears in the output, and \LMChatVThreePolicyEffectiveLeakN{} is a judge flag
the reviewers confirmed. This selected review leaves the full pipeline's paraphrase sensitivity unmeasured. A scan of
each history's surviving suffix for lexical contact with the deleted answer did
not explain the residual either: disclosure was, if anything, less frequent
among contact histories than among clean ones (Fisher two-sided
\(p=\LMChatSuffixFisherP\)). Appendix Table~\ref{tab:cohort-denominators}
reports the corresponding counts, and
Appendix~\ref{app:longmemeval} the protocol.

\section{Retained-answer matching on the decoded histories}\label{sec:retained-utility}
\input{RETAINED_UTILITY_INSERT}
\FloatBarrier

\section{Discussion}\label{sec:discussion}

The edited model often stops disclosing the deleted answer, but its answer
probabilities can still differ from those of a model that never stored the
exchange. This matters even when the edit suppresses the answer more
strongly. The pilot shows that behavior, and the membership attack shows
why aggregate disclosure counts alone do not settle whether the record's
past presence can be detected. Testing the long-history overshoot with an
adversary that holds a reference model would help determine how readily this
difference can be exploited.

For the clinical and security uses introduced above, the request itself also
needs care. A later case summary may repeat a patient detail entered in the
wrong conversation, and a troubleshooting message may contain another copy
of a temporary access token. Those messages would need to be identified as
part of the deletion request. Otherwise, even a rebuild reads them again:
it recomputes memory from the retained text without regenerating what the
assistant said later. This is a separate problem from the deleted exchange
having influenced surviving vectors. Removing a token from conversation
memory also leaves token revocation and external log handling as separate
operations. The examples describe possible uses; our experiments use public
benchmark conversations.

There is also a practical tradeoff in choosing \(\nu\). It determines the
coefficient cap and feasibility limit \(1-\nu\), while affecting the margin
set available to the decrement and the model's ability to recall records.
The frequent refit fallback at 4B means that a local operation cannot yet be
assumed to be a fast one. In the matched cost study, the FP32 proxy took about
the same update time as rebuilding the graft and substantially longer than
rebuilding the base model. Recomputing all solve points contributes to this
implementation's cost; these measurements neither establish a lower bound nor
measure an optimized decrement restricted to affected gates. Tuning the
configuration separately for 1B and 12B, and timing exact decrement and repack
on the same requests, would establish more clearly where the interface is useful.

Utility also depends on the comparator. Matching binary admission on six of
eight records did not ensure matching target probabilities. In the matched
cost study, the proxy's retained-target log probability was lower than the base
rebuild's and higher than the graft rebuild's. The former comparison includes
the architecture change: even before deletion, the graft's retained-target
score was 0.818 nats/token lower than the base model's on these probes. On the
separate decoded histories, edited retained-answer matching was 44/96 versus
46/96 for rebuilding under the broader matcher. The paired intervals allow
losses, so these findings do not establish preservation of general capability.
Repeated deletions and growth beyond sealed solve points were not evaluated
in this model installation.

Our disclosure measurements depend on what the screening process can
recognize. The judge is served under a provider name that can change between
calls, and human validation covers its flags and a control sample. The
review explains the selected outputs but leaves paraphrase sensitivity
across the full pipeline unmeasured. More extensive review would help assess
that limitation. Memory representation also affects the cost of a stronger
comparison: in native delta-attention systems, the tested receipt corrections
left a residual, while checkpoint replay reached the declared never-stored
reference at a cost that grew with the surviving suffix~\citep{kdaomission}.

\section{Conclusion}

We installed a support-vector gate in a frozen Gemma checkpoint so that a
deletion request can locate an exchange's stored rows, exclude them, and
check the resulting operation. The experiments show that this is possible
without further training at the tested 4B configuration. They also show
where the operation falls short: fewer records pass the admission checks at
other scales, and excluding a record's own rows can leave evidence that it
was once present. Comparing with memory rebuilt from the retained history
helps expose that remaining influence. The matched cost study found no
update-speed advantage for the current FP32 proxy, and retained-answer matching
remained below half in every tested condition. The interface makes a local
edit checkable, while these measurements identify its current practical limits.

\FloatBarrier
\label{main-text-end}

\section*{Ethics statement}
TOFU facts are fictitious and LongMemEval uses public benchmark conversations.
Figure~\ref{fig:leak-verbatim} contains selected fictitious TOFU completions
already included in the public manuscript. Full decoded response sets and
human-review packets are not included in the submission source package.
The experiments do not establish legal compliance or evaluate clinical use.

\section*{Reproducibility and data availability}
Code and supporting evidence are available at
\url{https://github.com/VyLabs-AI/gemmasv}.
The \texttt{journal\_evidence} directory contains the source-free journal
study supplement, including its file manifest and verification instructions.
The study artifacts pin Gemma revisions, dataset revisions, solver seed,
bandwidth rule, confirmation cohort, probes, and source-state hashes.
Generated macros bind LongMemEval values to source-free aggregates.
The cost study and offline reanalysis have separate protocols, manifests,
and source snapshots. The cost study's final audit checked all eight records,
48 arm outcomes, tokenized contexts and probes, and 529 snapshot files;
model-weight shards were not rehashed. No model calls were made during the
offline retained-answer reanalysis. The editable manuscript package contains
only manuscript dependencies and public aggregate figures, not model weights
or raw benchmark histories. The source-free Supplementary Material
(\texttt{gemmasv\_journal\_evidence\_v1.zip}) provides the study protocols,
per-record results, analyses, audit reports, and their provenance. Raw
benchmark histories, full generated responses, model weights, and private
review ledgers are not redistributed in that package. TOFU and LongMemEval
source data are available through their original benchmark distributions
\citep{tofu,longmemeval}. Access to original datasets and checkpoints remains
subject to their distributors' access and licensing terms. Local artifact paths in the provenance identify the original execution environment;
the repository above is the access point for the distributed evidence.

\section*{Funding}
This research received no funding.

\section*{Declaration of competing interests}
Vishwajith Ramesh is CEO and Founder of Vy Labs, Inc. He declares no other
competing financial or non-financial interests and no editorial role at
Neurocomputing.

\section*{Declaration of generative AI and AI-assisted technologies}

Language-model tools, including OpenAI Codex (GPT-6), assisted with literature
search, experimental design, code development, study execution and analysis,
figure preparation, and manuscript editing, including the cost comparison and
post hoc retained-answer analysis. AI assistance is not independent
human verification of the claims or artifacts.

%% file: COST_UTILITY_INSERT.tex
We ran a separately specified comparison on the eight fixed TOFU records
using the frozen 4B-PT checkpoint and the mass-preserving configuration applied at each solve point ($\nu=0.7$, window 1024, solver seed 0; no adapter or training).
Execution used the MPS backend on macOS 26.6.2 (arm64), Python 3.11.14,
PyTorch 2.12.1, and Transformers 5.12.1. The local model snapshot was
\texttt{cc012e0a6d}\allowbreak\texttt{0787b4adcc}\allowbreak\texttt{0fa2c4da74}\allowbreak\texttt{402494554d}.
The records were reused from the historical confirmation cohort; the new
execution protocol was frozen before inference, so this is not a new holdout.
All records were attempted without filtering by admission. The arms share
the original context and probes; each architecture's fresh reference
prefills the literal edited token sequence, which is shorter than the
historical padded never-stored control. Full target sequences are scored in log probability; within-architecture
first-token comparisons use full-vocabulary $\mathrm{KL}(p^{\rm rebuild}\Vert
p^{\rm arm})$. Cross-architecture KL uses the base rebuild as its reference
and includes architecture effects. These are output comparisons, not the
fixed-key solver certificate. The comparison includes the
ungrafted model with raw-record removal and cache rebuilding, the graft's
fresh rebuild, an FP32 masked-refit proxy, and diagnostic cache shifting.
The proxy implements logical exclusion through a drop mask while retaining
copied token IDs and K/V arrays. It recomputes gates for every eligible head at each stored solve point in every graft layer, including unaffected prefixes; it
does not restrict work to affected gates. Contextualized keys are reused
without language-model re-ingestion unless the feasibility precheck
triggers a full rebuild. Physical erasure from storage is not evaluated.
No exact-decrement or exact-certificate construction was timed.

Each arm used one warmup and three measured repetitions. We separately
timed deletion-state construction and four teacher-forced target probes
(three deleted fields and one retained field), with device synchronization.
Record lookup, tokenization, locating deletion positions, and constructing
the edited token list were outside the timer; model loading and shared
original prefill were separate. Query timing includes state cloning and
the existing sequential teacher-forced scorer. These are audit-workload
times, not production response latency. Table~\ref{tab:journal-bridge-cost}
reports absolute costs; the target scores are not decoded leakage rates.

\begin{table*}[t]
\centering\small
\caption{Completed journal bridge. Seconds are medians across the per-record median times; independently summarized columns need not add. The FP32 proxy is not an exact-decrement implementation, and cache shifting is diagnostic.}
\label{tab:journal-bridge-cost}
\begin{tabular}{lrrrr}
\toprule
Arm & Complete & Update (s) & Four-probe audit (s) & Update + audit (s) \\
\midrule
Base present control & 8/8 & 0.000 & 8.100 & 8.100 \\
Base literal rebuild & 8/8 & 0.701 & 8.080 & 8.770 \\
Graft present control & 8/8 & 0.000 & 9.041 & 9.041 \\
Graft literal rebuild & 8/8 & 24.882 & 8.905 & 33.134 \\
Graft FP32 masked refit & 8/8 & 25.141 & 11.904 & 36.886 \\
Graft cache-shift diagnostic & 8/8 & 0.051 & 9.045 & 9.098 \\
\bottomrule
\end{tabular}
\end{table*}

The FP32 proxy used a full-rebuild fallback for solve-point feasibility in 0/8 completed record evaluations. This is distinct from the historical per-head exact-decrement fallback rate.

\begin{table*}[t]
\centering\small
\caption{Paired FP32 masked-refit comparisons. Probability differences are mean target log probability per token, proxy minus reference. Cost ratios are proxy/reference, with values above one indicating a slower proxy. Brackets are 95\% percentile sensitivity intervals resampling four source blocks (10{,}000 draws), preserving both records in each block.}
\label{tab:journal-bridge-paired}
\begin{tabular}{lrr}
\toprule
Paired measure & Graft rebuild reference & Base rebuild reference \\
\midrule
Deleted target $\Delta$ LP/token & $0.0431\ [0.0046, 0.0816]$ & $-0.0093\ [-0.0519, 0.0187]$ \\
Retained target $\Delta$ LP/token & $0.3785\ [0.1798, 0.5772]$ & $-0.5240\ [-1.1366, -0.0628]$ \\
Update ratio & $1.0120\ [0.9882, 1.0367]$ & $35.7866\ [35.1790, 36.5365]$ \\
Update + audit ratio & $1.0800\ [1.0537, 1.1082]$ & $4.4274\ [4.1571, 4.6311]$ \\
\bottomrule
\end{tabular}
\end{table*}

This implementation showed no update-speed advantage over either rebuild comparator. The FP32 proxy's paired update ratio was 1.012 relative to rebuilding the graft and 35.79 relative to rebuilding the ungrafted base. Its update-plus-four-probe ratio was 1.080 and 4.43, respectively. These are geometric means of within-record ratios, not ratios of the table medians. The proxy's retained-target mean log-probability difference was -0.524 nats/token versus the base rebuild and +0.378 versus the graft rebuild; the base comparison includes architecture differences. The cache-shift diagnostic's retained-target difference was -5.045 nats/token versus graft rebuilding. Recomputing gates at every solve point is an implementation cost, not a fundamental complexity bound.

The cohort reuses four source blocks, whose independence is not established;
bootstrap intervals are descriptive sensitivity summaries, not evidence of
population equivalence or noninferiority. The base and graft ran sequentially,
and three timing repetitions are not three independent datasets. The
full report retains all per-record outcomes, target-level probability
differences, and failures. These measurements neither establish preservation
of general capability nor justify running an exact certificate on every
production deletion request.

%% file: RETAINED_UTILITY_INSERT.tex

We additionally compared retained-answer correctness using the existing decoded
follow-up: all 96 histories, nested three per target cluster across 32 clusters.
This was a post hoc reanalysis of source-free per-history matcher flags, with
aggregate outcomes already inspected before its analysis plan. It involved no
new generations or model calls. We retained both the legacy registered
casefold-substring endpoint and the existing broader deterministic matcher;
the latter is a secondary instrument, not a semantic-correctness judgment.
All 96 histories were scorable in the edited-policy, fresh-rebuild, and
record-present conditions. Repeated deterministic decodes supplied
reproducibility checks, not additional independent observations.

For each comparison, we averaged the paired binary correctness differences
within each cluster and then averaged the 32 clusters equally. We computed
95\% percentile intervals from 100{,}000 bootstrap resamples of target clusters,
keeping all three history variants together (NumPy \texttt{default\_rng},
seed 20260906). The all-history operational endpoint would count an unavailable
or unscorable response as incorrect; none occurred here. Table~\ref{tab:retained-offline-paired}
reports absolute matching counts and all three paired comparisons for both
endpoints. Its six intervals are descriptive and have no multiplicity correction.

\begin{table*}[t]
\centering
\small
\caption{Post hoc retained-answer matching from the existing decoded outputs.
Every absolute count has denominator 96, with no unscorable histories.
Differences and interval bounds are percentage points, first condition minus
second; positive differences favor the first condition. Intervals resample
the 32 target clusters. Matching the gold descriptor is a limited correctness
proxy; equality between two responses does not establish correctness.}
\label{tab:retained-offline-paired}
\begin{tabular}{lrrr}
\toprule
Existing matcher & Edited policy & Fresh rebuild & Record present \\
\midrule
Broader deterministic & 44/96 & 46/96 & 46/96 \\
Registered substring & 38/96 & 39/96 & 39/96 \\
\bottomrule
\end{tabular}

\medskip
\begin{tabular}{llrr}
\toprule
Existing matcher & Paired comparison & Difference & 95\% cluster interval \\
\midrule
Broader deterministic & Policy $-$ rebuild & $-2.08$ & $[-6.25,\ 2.08]$ \\
 & Policy $-$ present & $-2.08$ & $[-8.33,\ 4.17]$ \\
 & Rebuild $-$ present & $0.00$ & $[-6.25,\ 6.25]$ \\
\midrule
Registered substring & Policy $-$ rebuild & $-1.04$ & $[-4.17,\ 2.08]$ \\
 & Policy $-$ present & $-1.04$ & $[-5.21,\ 3.12]$ \\
 & Rebuild $-$ present & $0.00$ & $[-5.21,\ 5.21]$ \\
\bottomrule
\end{tabular}
\end{table*}

The edited policy's retained matching rate was slightly below the fresh
rebuild's point estimate under both instruments. These differences do not
establish equivalence or noninferiority: no corresponding margin was specified,
the intervals permit losses, and absolute matching remained below 50\% in
every condition. This analysis supplies paired retained-answer accounting on
the existing outputs; it does not provide the still-unavailable teacher-forced
forced-choice margins, gold-token ranks, or continuation KL, and it does not
establish preservation of general model capability.

%% file: appendix_gemma.tex
\section{Audit references and denominators}\label{app:contracts}

The experiments count several different things, from admitted records to
generated answers. Table~\ref{tab:cohort-denominators} keeps those units and
their denominators together. In particular, the human reviewers saw the
judge's flags and a control sample; that selected set cannot supply the
denominator for a condition's disclosure rate.

\begin{table*}[htbp]
\centering
\caption{Cohorts and denominators for every analysis in the paper. Unlike
units are named separately instead of being collapsed into one \(N\).
Qualification means strict recall admission for the Gemma and LongMemEval
cohorts and condition completion for the decoded outputs; the column
vocabulary matches the denominator table of the delta-attention study~\citep{kdaomission}.}
\label{tab:cohort-denominators}
\scriptsize
\setlength{\tabcolsep}{3pt}
\begin{tabular}{@{}p{0.18\textwidth}p{0.20\textwidth}p{0.23\textwidth}p{0.30\textwidth}@{}}
\toprule
surface & audited unit & attempted & qualified \\
\midrule
Gemma confirmation & record; certificate probe &
8 records per scale & admission \(3/8\) 1B, \(6/8\) 4B, \(5/8\) 12B;
32 probes at each certified scale \\
Historical LongMemEval & history; probe &
\(n=16\) histories & 10 jointly admitted; 32 certificate probes \\
Decoded LongMemEval & cluster; history; decoded output &
\(K=\LMChatVThreeK\) clusters; \(n=\LMChatVThreeN\) histories;
\LMChatVThreePopulationN{} decoded outputs over four conditions &
all \LMChatVThreeCompletedN{} histories completed; the matcher flagged
\LMChatVThreeMatcherPositiveN{} outputs and cleared
\LMChatVThreeMatcherCleanN{}, all of which the judge read \\
New cost bridge & record; timing repetition &
8 reused records in 4 source blocks; 6 arms each &
48/48 arms completed; 1 warmup and 3 timing repetitions per arm;
0/8 proxy record-level solve point-precheck rebuilds \\
Retained-answer reanalysis & target cluster; history &
32 clusters; 96 existing histories &
all 96 scorable in each of 3 conditions; two matcher endpoints;
no new generations \\
Broad LiRA & record; positive/negative draw &
20 records & 16 admitted; 512 draws per side \\
Human review & decoded output &
\LMChatVThreeHumanFlagN{} outputs the judge flagged and
\LMChatVThreeHumanControlN{} it cleared, chosen by hash &
flagged: \LMChatVThreeHumanFlagLeakN{} disclosures,
\LMChatVThreeHumanFlagNoLeakN{} not, \LMChatVThreeHumanFlagAmbiguousN{}
ambiguous; controls: \LMChatVThreeHumanControlLeakN{} disclosures in
\LMChatVThreeHumanControlN{}; second reviewer agreed on all
\LMChatVThreeHumanSecondaryTripleN{} texts read \\
\bottomrule
\end{tabular}
\end{table*}

\FloatBarrier

\section{Gemma protocols and complete outcomes}\label{app:gemma}

The evaluated checkpoints are
\texttt{google/gemma-3-\{1b,4b,12b\}-pt}. Only global layers are replaced; all
pretrained parameters remain fixed. The configuration uses \(\nu=0.7\), chunk length
128, solver seed 0, a deterministic 256-key median bandwidth, prefix-mass
preservation, and \(C_b=1/(\nu n_b)\) at each sealed prefix solve point.

The gate also preserves differences in the model's original attention.
For an unequal-share arithmetic example, let the older rows have
$a=(0.2,0.1,0.1)$ and $\alpha=(0.5,0,0.5)$. Their products are
$(0.1,0,0.05)$; restoring the older share $0.4$ gives
$\widetilde a=(4/15,0,2/15)$, while recent context keeps $0.6$.
The normalization in Equation~\ref{eq:attention-gate} thus retains the
differences between the original attention weights even when the nonzero
gate coefficients are equal.

The eight-record confirmation cohort was frozen after configuration selection. Strict admission
requires three target fields, one retained neighbor, and feasibility at every
affected solve point. Every outcome remains in the denominator. Quality uses the
same 400 packed WikiText-103 blocks at all scales.

\subsection{Scale sensitivity}\label{app:gemma-scale-sensitivity}

The scale outcomes are in main-text Table~\ref{tab:scale}.

\begin{figure}[!t]
\centering
\includegraphics[alt={Across Gemma 1B, 4B, and 12B checkpoints, only 4B preserves base strict admission with low paired perplexity overhead.},width=\linewidth]{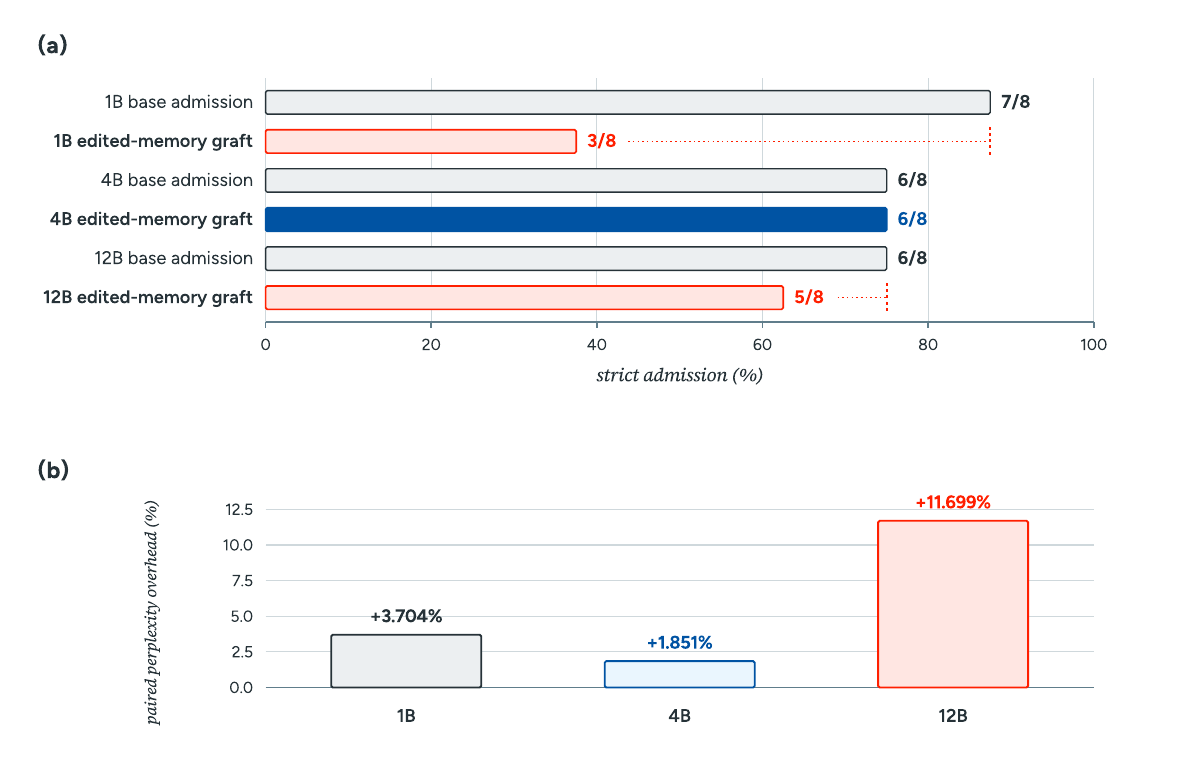}
\caption{Admission and quality cost of the frozen configuration at three scales. (a)
Strict admission, where 4B is the only tested scale that preserves the base
model's eligibility. (b) Paired perplexity overhead, where 12B has the largest
measured quality cost. A dashed marker indicates that the graft no longer
matches its own base admission. The three checkpoints are sensitivity outcomes
and were not selected by any model-selection procedure.}
\label{fig:audit-boundaries}
\end{figure}

\FloatBarrier

We checked both agreement with the refit and completion of the decrement at
the two certified scales. At 1B, the median and maximum KL against the refit on retained keys are
\(1.86\times10^{-13}/1.25\times10^{-10}\); at 4B they are
\(2.36\times10^{-12}/6.48\times10^{-11}\). No probe exceeds \(10^{-6}\).
The reverse decrement failed its checks in \(396/640\) 1B and
\(2{,}020/3{,}040\) 4B attempted affected solves. Refitting completed each
of those updates, so every deletion still reached its reference. These
denominators include affected solves only; solve points before the deleted
position do not need an update. Scoring the already-constructed exact, refit, and decay states
takes 561.5 s for 32 probes at 1B and 2,354.6 s at 4B; these are the costs of
scoring the audit and do not measure deletion latency.

\subsection{Behavioral attack details}

At \(k=1\), exact-phrase leakage is \(0.28\%\) edited and \(0.42\%\)
never-stored. At \(k=200\), each condition leaks on \(3/18\) field queries
(\(16.7\%\), the cohort rate), with paired record-bootstrap difference
\(0.0\,[0.0,0.0]\). Present and prompt-only are \(18/18\) (\(100\%\)), and decay
is \(5/18\) (\(27.8\%\)). To check reproducibility, we repeated the sampling
arm with the same seeds, decoding settings, and batch size. It reproduced every
\(k=200\) endpoint exactly (\(18/18\), \(18/18\), \(3/18\), \(3/18\),
and \(5/18\) for present, prompt-only, edited, never stored, and decay) and
the \(k=1\) rates to within \(0.02\) percentage points, while individual
sampled completions differed. Thus the per-query endpoints reproduced on
this hardware even though the per-sample counts were not bit-exact.
Elicitation, related-data updates, and LiRA are described in the released
attack aggregate.

For LiRA AUC intervals, we resampled records with replacement and took
percentile endpoints over \(10^4\) draws: six-record
cohort \(0.534\) \([0.513, 0.601]\), broad cohort \(0.517\)
\([0.516, 0.539]\); AUCs computed separately for each record span \(0.47\)--\(0.76\) and
\(0.53\)--\(0.73\). The broad LiRA policy includes
\(16/512\) edited test deletions routed to raw repack. Replacing those scores
with arbitrary values moves the AUC anywhere in \([0.486,0.549]\); we report
that range as a sensitivity check, distinct from a confidence interval.
The frozen interval code sorts the 10,000 pooled AUCs and selects ordered
entries 251 and 9,750 (one-based), yielding the asymmetric broad interval
[0.5157394409179688, 0.5387916564941406]. It also reports the upper central
AUC computed separately for each record for even sample sizes. Table~\ref{tab:lira-results} keeps
these conventions and rounds only for display.

\section{LongMemEval protocols and complete counts}\label{app:longmemeval}

Section~\ref{sec:longmemeval} reports the findings of the two LongMemEval
studies. This appendix gives their protocols and the complete counts behind
each reported rate, beginning with the selected examples and then describing
the two cohorts.

\subsection{Selected decoded examples and their limits}\label{app:decoded-cases}

The mortgage, screwdriver, and tennis examples are post hoc illustrations
from the completed 96-history study, not additional experiments. Each target
has three nested suffix variants assembled from the pinned public LongMemEval
oracle, with the same target and retained control. The protocol deletes the
selected latest user--assistant round, preserves earlier evidence literally,
and leaves later retained turns unchanged. A fresh raw-history reference
re-encodes those retained inputs; it does not regenerate them. For each probe,
the model receives a registered user turn containing \texttt{Current Date:},
the official question, and ``Reply with only the remembered value, without
explanation.'' A model turn then begins. Repeated deterministic decodes are
reproducibility checks, not independent samples. Solver checks target the
fixed-$C$ problem on the retained keys; no separate refit response was decoded.

\paragraph{Mortgage update.}
The official target question is quoted in Section~\ref{sec:mortgage-case},
with query date December 9, 2023. The owned November 30 round occupies token
positions $[876,1233)$ in each assembled history. The earlier August 11 round
mentions pre-approval for \$350,000; the removed round mentions \$400,000.
The retained question, dated May 27, 2023, is ``I wanted to follow up on our
previous conversation about YouTube videos for workplace posture. Can you
remind me of the Mayo Clinic video you recommended?'' The saved benchmark gold
contains the video title ``How to Sit Properly at a Desk to Avoid Back Pain''
and URL \url{https://www.youtube.com/watch?v=UfOvNlX9Hh0}.

\begin{table}[ht]
\centering\small
\caption{All three mortgage variants. Entries are saved generated answers to
the target query; the video URL is present under all four conditions in every
variant. Variant 0's edited answer also includes the video title.}
\label{tab:mortgage-variants}
\begin{tabular}{@{}rrrrlll@{}}
\toprule
variant & tokens & after edit span & present & prompt & local edit & rebuild\\
\midrule
0 & 2,596 & 1,363 & \$400,000 & \$400,000 & \$350,000 & \$350,000\\
1 & 2,764 & 1,531 & \$400,000 & \$400,000 & \$350,000 & \$350,000\\
2 & 2,941 & 1,708 & \$400,000 & \$400,000 & \$350,000 & \$350,000\\
\bottomrule
\end{tabular}
\end{table}
The local policy used refit fallback in 480/560, 508/600, and 534/640 affected
head/solve-point solves, respectively. The high use of refitting leaves the speed advantage of a
local edit unestablished. The retained URL matches the broader endpoint in
all variants, although the registered whole-answer substring endpoint does
not match the complete gold sentence. Aggregate retained matching is
reported in Section~\ref{sec:mortgage-case}.

\paragraph{A local edit and rebuild can answer differently.}
For the screwdriver target, the earlier exchange says the small screwdriver
was misplaced. The latest user says ``I actually have a spare screwdriver'';
the assistant responds ``Spare screwdriver: Ah, excellent!'' [excerpts].
The deleted user--assistant round occupies $[1189,1594)$. The official query
is ``Do I have a spare screwdriver for opening up my laptop?''
Table~\ref{tab:screwdriver-variants} preserves the suffix-dependent outcome.
All conditions return ``38'' to an unrelated numerical control whose saved
gold is ``38 subjects''; neither existing retained matcher marks those answers
correct. The reported labels follow those registered matchers.

\begin{table}[ht]
\centering\small
\caption{All three screwdriver variants. Editing and rebuilding disagree in
variant 0; the other two rebuilt answers also contain the target ``Yes''.}
\label{tab:screwdriver-variants}
\begin{tabular}{@{}rrllll@{}}
\toprule
variant & tokens & present & prompt & local edit & rebuild\\
\midrule
0 & 3,160 & Yes & No & Yes & No\\
1 & 3,290 & Spare screwdriver & Yes & Yes & Yes\\
2 & 3,038 & Yes & No. & Yes & Yes\\
\bottomrule
\end{tabular}
\end{table}
The local policy used refit fallback in 538/600, 564/640, and 510/560 affected
solves. These examples show a difference in decoded behavior; they do not
identify which surviving row or later turn caused it.

\paragraph{A harder question can expose an endpoint limitation.}
The tennis target asks both previous and current frequency. Its full gold
answer says previously every week on Sunday and currently every other week
on Sunday. Across histories of 2,339, 2,356, and 2,674 tokens, present and
prompt-only conditions generate only ``Every other week'', whereas editing
and rebuilding generate ``Weekly''. The retained answer is
\texttt{MusicTheory.net} throughout. Neither partial target answer matches
the full registered gold, so matcher-negative output alone does not establish
successful forgetting. The frozen counts retain the registered matcher
labels for these incomplete answers.

The source question IDs are \texttt{852ce960} (mortgage),
\texttt{42ec0761} (screwdriver), and \texttt{f685340e} (tennis).
The source-free cohort manifest fixes all three history IDs and token spans
per target. The response audit connects the saved outputs to those histories,
and a local review manifest records their artifact paths and hashes.

\subsection{Bicycle and Speyer case: teacher-forced scores}\label{app:bicycle-case}

The bicycle example in the introduction comes from the 16-history pilot.
Its 3,103-token assembled history contains an earlier three-bike discussion;
the deleted round reports the new hybrid bike, updates the count to four,
and includes the assistant's echo. The retained exchange supplies the Speyer
phone number. We use teacher-forced scoring to see how deletion changes the
probabilities of the saved answers: the model receives each saved answer
token as it proceeds, and we record its probability. For ``How many bikes do I currently own?'',
the one-token gold ``4'' moves from rank 2 to rank 28 after editing.
The 22-token phone answer keeps rank 1 at its first token; its mean log
probability changes from $-2.31560\times10^{-5}$ to
$-4.64238\times10^{-5}$ nats per answer token. Because the previous gold
tokens are supplied during scoring, the phone-number result does not tell
us whether the model would generate the complete number on its own.
The maximum local output KL is $4.32788\times10^{-17}$ nats across two probes,
in the direction from the edited policy to the fixed-$C$ refit. The bicycle probe's distinct
KL from the raw rebuild to the edited policy is $0.00176439$ nats over the first-token vocabulary.
Of 560 affected solves, 522 used exact-refit fallback and 38 completed
incrementally. The record ID ends in \texttt{c8276e265e3db489c090cdcc}.

\LMChatMainResultText

\LMChatAppendixText

\input{longmemeval_chat_v3_decoded_followup}

%% file: longmemeval_chat_v3_decoded_followup.tex

\subsection{Decoded follow-up: cohort and decoding}\label{app:longmemeval-cohort}

The follow-up tests what the model actually generates on
\(K=\LMChatVThreeK\) target clusters and \(n=\LMChatVThreeN\) histories,
separate from the pilot above (Appendix Table~\ref{tab:cohort-denominators}).
We froze the \LMChatVThreeK{} clusters first and packed three histories per
cluster independently. Every history completed under four conditions: record
present, local deletion, rebuilding from the retained history, and a
prompt-only instruction to forget. This produced
\LMChatVThreePopulationN{} decoded outputs. We kept histories even when the
model failed to recall the record, so every rate below uses all
\LMChatVThreeN{} histories. Decoding each probe greedily twice produced
identical text and token identifiers, checking deterministic reproducibility
rather than variation across sampled answers.

\subsection{Screening for disclosure: a matcher, a judge, and two reviewers}\label{app:longmemeval-screening}

We screened the \LMChatVThreePopulationN{} decoded outputs in three stages.
First, a deterministic matcher compared each output with the source answer
under five checks applied together: a casefolded substring test, a test after
punctuation and articles are stripped, a numeric and unit test, a URL test,
and a named-entity test. An output that passed any of them counts as a
disclosure. There were \LMChatVThreeMatcherPositiveN{} of these, and we did
not review them further. Second, a language model judge
(GPT-5.6) read every one of the \LMChatVThreeMatcherCleanN{} outputs the
matcher had cleared, seeing the question, the reference answer, and the
candidate response but neither the condition nor the matcher's verdict, and
returned one of three labels: disclosure, no disclosure, or ambiguous. Every
request returned a valid response. Because the provider serves the judge
under a name that can change between calls, the local audit retains the
requested and returned model names along with the frozen prompt, response
schema, and every request and response.

Third, two reviewers read a sample of the outputs, blinded to condition.
The sample held the \LMChatVThreeHumanFlagN{}
outputs the judge had flagged and \LMChatVThreeHumanControlN{} outputs it had
cleared, the latter chosen by hash so that the choice could not depend on
their content (\LMChatVThreeHumanReviewN{} outputs in all). The first reviewer
labeled every output in the sample, and the second read the flagged outputs
the first had not labeled as disclosures (\LMChatVThreeHumanSecondaryTripleN{}
distinct texts) and agreed with the first on each. The
\LMChatVThreeHumanReviewN{} outputs contain only
\LMChatVThreeHumanUniqueTripleN{} distinct question, answer, and response
triples, because the same response text recurs across conditions. We count
outputs below, but the duplicate texts are not independent judgments.

Of the \LMChatVThreeHumanFlagN{} flagged outputs, the reviewers labeled
\LMChatVThreeHumanFlagLeakN{} as disclosures,
\LMChatVThreeHumanFlagNoLeakN{} as no disclosure, and
\LMChatVThreeHumanFlagAmbiguousN{} as ambiguous, and they labeled all
\LMChatVThreeHumanControlN{} controls as no disclosure. Every confirmed
disclosure was a form of the answer that the matcher had failed to recognize:
\LMChatVThreeHumanNormalizationN{} were normalization failures and
\LMChatVThreeHumanAliasMorphologyN{} were an alias or an inflected form of the
answer, and none was a paraphrase. Most reviewed outputs were selected
because the judge had flagged them. The review therefore characterizes those
flags and a control sample, rather than estimating how often each condition
discloses. The \LMChatVThreeJudgeAmbiguousN{} outputs the judge itself marked
ambiguous were not reviewed.

\subsection{Disclosure rates and the paired comparison}\label{app:longmemeval-rates}

For the endpoint an output counts as a disclosure when the matcher flagged it
or the reviewers labeled it one; for flagged outputs the reviewers did not
read, the judge's label stands. Ambiguous labels give a lower and an upper
count. The reviewed labels changed the aggregate counts as follows.
After review, the \LMChatVThreeMatcherCleanN{} outputs the matcher cleared hold
\LMChatVThreeEffectiveMatcherCleanLeakN{} disclosures, \LMChatVThreeEffectiveMatcherCleanNoLeakN{} non-disclosures, and
\LMChatVThreeEffectiveMatcherCleanAmbiguousN{} ambiguous outputs, against the
judge's \LMChatVThreeJudgeMissN{}, \LMChatVThreeJudgeNoLeakN{}, and
\LMChatVThreeJudgeAmbiguousN{} before review. Per history, the edited policy
disclosed the deleted answer in
\(\LMChatVThreePolicyLeakLowerN/\LMChatVThreeN\), the never-stored rebuild in
\(\LMChatVThreeRebuildLeakLowerN/\LMChatVThreeN\), the record-present run in
\LMChatVThreePresentLeakLowerN{}--\LMChatVThreePresentLeakUpperN{} of
\LMChatVThreeN{}, and the prompt-only instruction in
\LMChatVThreePromptLeakLowerN{}--\LMChatVThreePromptLeakUpperN{}. The paired
edited-minus-rebuild difference is
\(+\LMChatVThreeEditedMinusFreshLowerN/\LMChatVThreeN=+\LMChatVThreeEditedMinusFreshLowerPoints\)
percentage points, with a \(95\%\) interval of
\([\LMChatVThreeEditedMinusFreshLowerCILowerPoints,
\LMChatVThreeEditedMinusFreshLowerCIUpperPoints]\) (a bootstrap over the
\LMChatVThreeK{} clusters), and the prompt-minus-edited difference is at least
\LMChatVThreePromptMinusEditedLowerPoints{} points (interval
\([\LMChatVThreePromptMinusEditedLowerCILowerPoints,
\LMChatVThreePromptMinusEditedLowerCIUpperPoints]\)). These
intervals hold the labels fixed and do not include uncertainty from selecting
the control outputs or from the reviewers' judgments. The \(+2/96\)
difference does not resolve equivalence, because we set no equivalence margin
in advance. Against margins chosen afterwards, the upper end of the interval,
\LMChatVThreeEditedMinusFreshLowerCIUpperPoints{} points, misses a 5-point
margin and meets a 10-point one. The counts put the edit well below the
prompt instruction and within a few points of rebuilding on this probe,
with disclosures remaining under both procedures.

\subsection{Matching responses and correct responses}\label{app:longmemeval-correctness}

The edited policy returned exactly the same target response as the
never-stored rebuild in \LMChatVThreeTargetPolicyRebuildExactN{} of
\LMChatVThreeN{} histories, and the same response under at least one of the
matcher's normalizations in \LMChatVThreeTargetPolicyRebuildNormalizedN{}.
To assess retained memory, we also compared responses with their saved
answers. On the retained probe, the answer was correct under the matcher in
\(\LMChatVThreeRetainedPresentCorrectN/\LMChatVThreeN\) histories with the
record present, \(\LMChatVThreeRetainedRebuildCorrectN/\LMChatVThreeN\) after
the rebuild, and \(\LMChatVThreeRetainedPolicyCorrectN/\LMChatVThreeN\) under
the edited policy, and the edited policy's retained response matched the
record-present response under some normalization in
\(\LMChatVThreeRetainedPolicyPresentNormalizedN/\LMChatVThreeN\). This comparison matters because the edit and rebuild can agree on an
incorrect answer.

Two histories illustrate why response agreement and correctness need
separate measurements. We chose one from each suffix stratum by SHA-256 rank
before reading any response and kept every condition for each. In the history
whose suffix had no lexical contact with the deleted answer, the gold target
was ``38,'' but every condition answered ``37,'' including the record-present
control. The retained answer, ``Three,'' was stable throughout. Deletion
cannot uniquely explain a target error that was already present in the
control. In the history with lexical contact, the record-present and
prompt-only conditions answered ``5,'' while the rebuild and edit answered
``Four.'' Here the edit tracked the rebuild, but every condition missed the
paired retained restaurant. Both histories remain in the original cohort
and in the rates reported above.

\subsection{The frozen suffix}\label{app:longmemeval-suffix}

We examined the surviving suffix because later text can carry information
from a deleted exchange back into a rebuild. Copies can appear as an
assistant's echo, a derived value, or a summary; the delta-attention study
gives a worked example~\citep{kdaomission}. We scanned the turns after each
deleted record for lexical contact with the target answer and found it in
\LMChatVThreeSuffixContaminatedHistoryN{} of \LMChatVThreeN{} histories,
spanning \LMChatVThreeSuffixContaminatedClusterN{} of \LMChatVThreeK{}
clusters. The suffix was assembled from independent support sources and was
never generated downstream of the target, so contact marks possible
contamination and does not establish causal dependence. The scanner catches
explicit restatements and some normalized forms, generally misses derived
traces and summaries, and can match short or numeric strings. The detected count is neither a lower
nor an upper bound on contamination.

We assessed whether lexical contact was associated with the edited policy's residual
disclosures (the \LMChatVThreePolicyLeakLowerN{} of \LMChatVThreeN{} in
Section~\ref{sec:longmemeval}). Among the
\LMChatSuffixContactHistoryN{} histories with contact, \LMChatSuffixContactDisclosureN{} disclosed and
\LMChatSuffixContactNoDisclosureN{} did not; among the
\LMChatSuffixCleanHistoryN{} without contact, \LMChatSuffixCleanDisclosureN{}
disclosed and \LMChatSuffixCleanNoDisclosureN{} did not, for disclosure rates
of \(\LMChatSuffixContactRiskPct\%\) and \(\LMChatSuffixCleanRiskPct\%\)
(Fisher's exact test, two-sided, \(p=\LMChatSuffixFisherP\)). Disclosure was less frequent in the lexical-contact stratum. The
matcher missed \LMChatSuffixHumanConfirmedEditedMissN{} edited-policy output
that the judge then flagged, and the reviewers confirmed it, so the
cross-tabulation is unchanged from the judge's labels. Lexical contact
therefore does not explain the residual disclosure in this study. A broader
deletion request could also ask for later turns to be regenerated if they
were caused by the deleted record. Our frozen-history comparison does not
test that operation; it is discussed in the delta-attention
study~\citep{kdaomission}.